\documentclass[graybox]{svmult}

\usepackage{mathptmx}       
\usepackage{helvet}         
\usepackage{courier}        
\usepackage{type1cm}        
\usepackage{makeidx}         
\usepackage{graphicx}        
\usepackage{multicol}        
\usepackage[bottom]{footmisc}
\usepackage{multirow}
\usepackage{booktabs}
\usepackage{subfig}
\usepackage{amsmath,amssymb}
\usepackage{bm}

\newcommand{\scal}[1]{\mathit{#1}}
\newcommand{\vect}[1]{\mathbf{#1}}
\newcommand{\matr}[1]{\mathbf{#1}}
\newcommand{\const}[1]{\mathit{#1}}

\newcommand{\revise}[1]{\textcolor{black}{#1}}

\newcommand{\mathbbm}[1]{\text{\usefont{U}{bbm}{m}{n}#1}} 

\makeindex             

\begin{document}

\title*{Zero-Shot Crowd Behaviour Recognition}
\author{Xun Xu, Shaogang Gong and Timothy M. Hospedales}
\institute{Xun Xu \at Queen Mary University of London, London, UK \email{xun.xu@qmul.ac.uk}
\and Shaogang Gong \at Queen Mary University of London, London, UK \email{s.gong@qmul.ac.uk}
\and Timothy Hospedales \at Queen Mary University of London, London, UK \email{t.hospedales@qmul.ac.uk}}
%
%
\maketitle

\abstract{Understanding crowd behaviour in video is challenging for
  computer vision. There have been increasing attempts on modelling
  crowded scenes by introducing ever larger property ontologies
  (attributes) and annotating ever larger training datasets. However,
  in contrast to still images, manually annotating video attributes
  needs to consider spatio-temporal evolution which is inherently much
  harder and more costly.  Critically, the most interesting crowd
  behaviours captured in surveillance videos (e.g. street fighting,
  flash mobs) are either rare, thus have few examples for model
  training, or unseen previously. Existing crowd analysis techniques
  are not readily scalable to recognise novel (unseen) crowd
  behaviours. To address this problem, we investigate and develop
  methods for recognising visual crowd behavioural attributes {\em
    without} any training samples, i.e. {\em zero-shot learning}
  crowd behaviour recognition. To that end, we relax the common
  assumption that each individual crowd video instance is only
  associated with a single crowd attribute. Instead, our model learns
  to jointly recognise multiple crowd behavioural attributes in each
  video instance by exploring multi-attribute co-occurrence as
  contextual knowledge for optimising individual crowd attribute
  recognition. Joint multi-label attribute prediction in zero-shot
  learning is inherently non-trivial because co-occurrence statistics
  does not exist for unseen attributes. To solve this problem, we
  learn to predict cross-attribute co-occurrence from both online text
  corpus and multi-label annotation of videos with known attributes. Our experiments
  show that this approach to modelling multi-attribute context not
  only improves zero-shot crowd behaviour recognition on the
  \textit{WWW} crowd video dataset, but also generalises to novel
  behaviour (violence) detection cross-domain in the
  \textit{Violence Flow} video dataset.
}

\section{Introduction}
\begin{figure}[t]
\centering
\subfloat[Thumbnails of {\em WWW} crowd video dataset and attributes\cite{Shao2015}]{\includegraphics[width=0.47\linewidth]{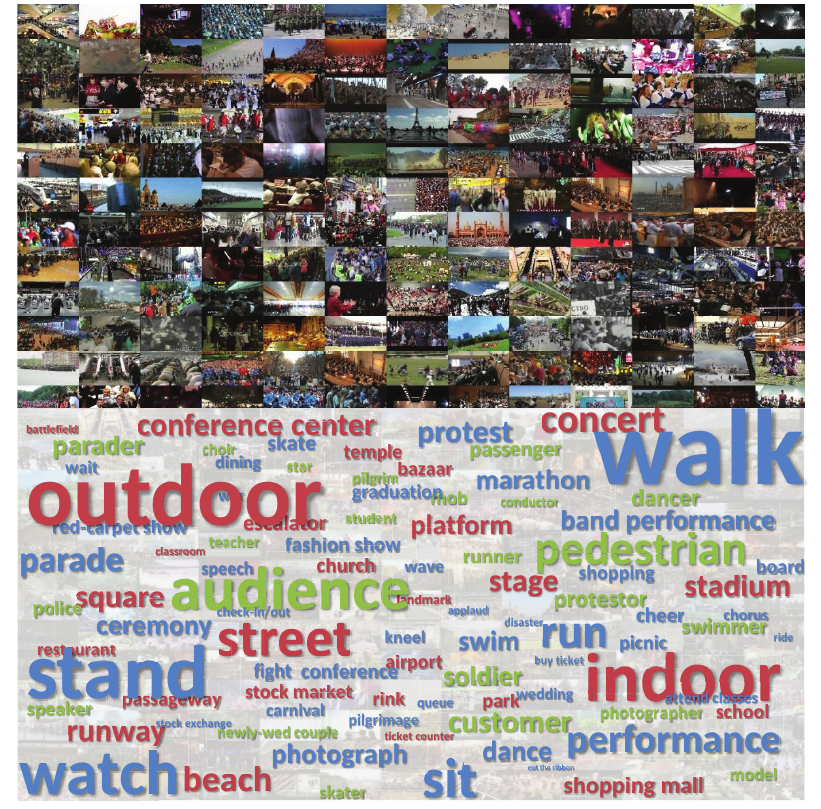}}
\subfloat[Frequencies of attributes]{\includegraphics[width = 0.52\linewidth]{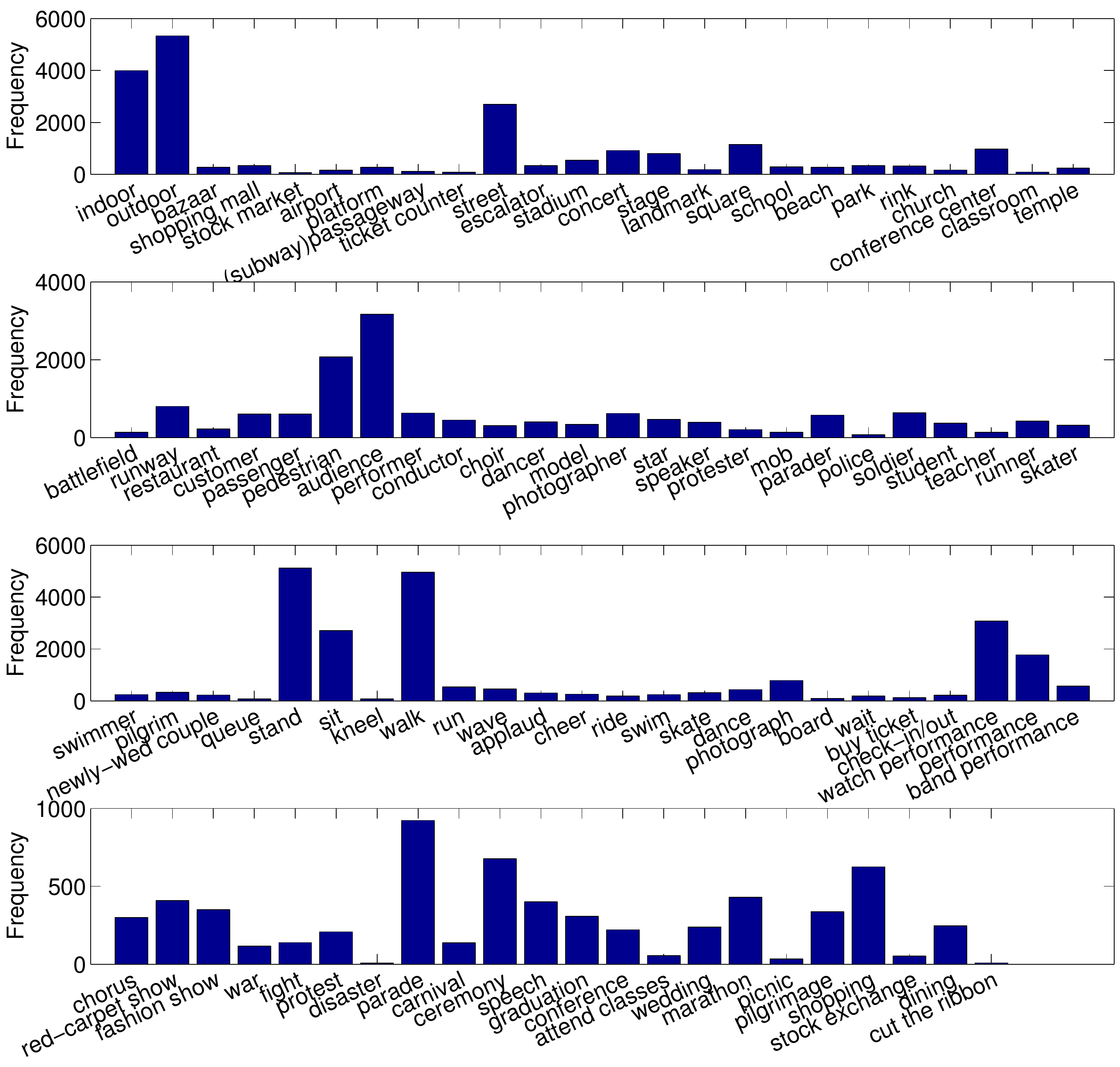}}\\
\caption{A thumbnail visualisation and a summary on the popularities of all 94
  attributes in the {\em WWW} crowd video dataset \cite{Shao2015}.}\label{fig:FreqAttrAll}
\end{figure}

Crowd behaviour analysis is important in video surveillance for public
security and safety. It has drawn increasing attention in computer
vision research over the past decade
\cite{xu2016discovery,journals/ijcv/VaradarajanEO13,wang2009unsupervised,rodriguez2011data,shao2014scene,Shao2015}. Most
existing methods employ a video analysis processing pipeline that
includes: Crowd scene representation
\cite{xu2016discovery,journals/ijcv/VaradarajanEO13,wang2009unsupervised,rodriguez2011data},
definition and annotation of crowd behavioural attributes for detection
and classification, and learning discriminative recognition models
from labelled data \cite{shao2014scene,Shao2015}. However, this
conventional pipeline is limited for scaling up to recognising ever
increasing number of behaviour types of interest, particularly
for recognising crowd behaviours of no training examples in a new
environment. Firstly, conventional methods rely on exhaustively
annotating examples of every crowd attribute of interest
\cite{Shao2015}. This is often implausible nor scalable due to the
complexity and the cost of annotating crowd \emph{videos} which
requires spatio-temporal localisation. Secondly, many crowd attributes
may all appear simultaneously in a single video instance,
e.g. ``\textit{outdoor}'', ``\textit{parade}'', and
``\textit{fight}''. To achieve \emph{multi-label} annotation
consistently, it is significantly more challenging and costly than
conventional single-label multi-class annotation. Moreover, the most
interesting crowd behaviours often occur rarely, or have never
occurred previously in a given scene. For example, crowd attributes
such as ``\textit{mob}'', ``\textit{police}'', ``\textit{fight}'' and
``\textit{disaster}'' are rare in the {\em WWW} crowd video dataset, both
relative to others and in absolute numbers (see
Fig.~\ref{fig:FreqAttrAll}). Given that such attributes have few or no
training samples, it is hard to learn a model capable of detecting and
recognising them using the conventional supervised learning based crowd analysis approach. 

In this chapter, we investigate and develop methods for zero-shot
learning (ZSL) \cite{Lampert2009} based crowd behaviour
recognition. We want to learn a generalisable model on well annotated common crowd attributes. Once learned, the model can then be deployed to recognise novel (unseen) crowd behaviours or attributes of interest without any annotated training samples. 
The ZSL approach is mostly exploited for
object image recognition: A regressor\cite{Socher2013} or classifier\cite{Lampert2009} is commonly learned on known categories to map a image's visual feature to the continuous semantic representation of corresponding category or the discrete human-labelled semantic attributes. Then it is deployed to project unlabelled images into the same semantic space for recognizing previously unseen object categories
\cite{Lampert2009,Socher2013,frome2013devise,akata2015outputEmbedding}.  
There have also been recent attempts on ZSL recognition of
single-label human actions in video instances
\cite{Xu_SES_ICIP15,AlexiouXG_ICIP16} where similar pipeline is adopted. However, for ZSL crowd
behaviour recognition, there are two open challenges. First, crowd
videos contain significantly more complex and cluttered scenes making
accurate and consistent interpretation of crowd behavioural attributes
in the absence of training data very challenging. Second, crowd
scene videos are inherently multi-labelled. That is, there are almost
always multiple attributes concurrently exist in each crowd video instance. The most interesting ones are often related to other non-interesting attributes. Thus we wish to infer these interesting attributes/behaviours from the detection of non-interesting but more readily available attributes. However this has not been sufficiently studied in crowd behaviour recognition, not to mention in the context of zero-shot learning.

It has been shown that in a {\em fully supervised setting}, exploring
co-occurrence of multi-labels in a common context can improve the
recognition of each individual label
\cite{zhang2014review,ghamrawi2005collective,conf/uai/0013ZG14}. For example, the behavioural attribute
``\textit{protest}'' \cite{Shao2015} is
more likely to occur in ``\textit{outdoor}'' rather than
``\textit{indoor}''. Therefore, recognising the indoor/outdoor
attribute in video can help to predict more accurately the ``\textit{protest}'' behaviour.  However,
it is not only unclear how, but also non-trivial, to extend this idea to the
ZSL setting.  
For instance, predicting a previously unseen behaviour
``\textit{violence}'' in a different domain \cite{hassner2012violent} would be much harder than the prediction of ``\textit{protest}''. As it is unseen, it is
impossible to leverage the co-occurrence here as we have no {\em a priori}
annotated data to learn their co-occurring context. The
problem addressed in this chapter is on how to explore contextual
co-occurrence among multiple known crowd behavioural attributes in order to
facilitate the prediction of an unseen behavioural attribute, likely
in a different domain.

\begin{figure}[t]
\centering
\includegraphics[width=0.99\linewidth]{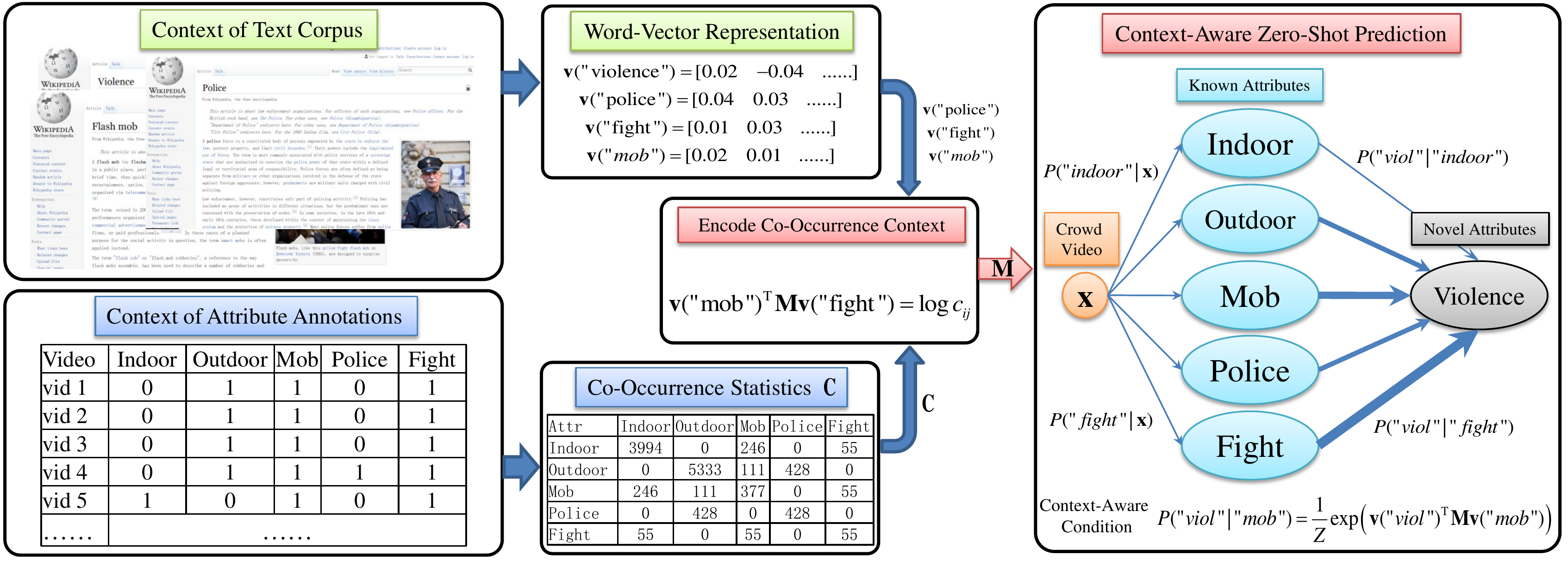}
\caption{In model training, we learn word-vector representations
  of training attributes from an external text corpus (context of text corpus), and their
  visual co-occurrence from the training video annotations (context of attribute annotations). A bilinear
  mapping $\matr{M}$ between pairs of word vectors is trained to
  predict the log visual co-occurrence statistics $\log
  c_{ij}$. Visual co-occurrence probabilities can be estimated for any
  pairs of known or novel (unseen) attributes. To enable the
  prediction of a novel attribute ``\textit{violence}'' using the
  context of known attributes, we first learn a recogniser for each
  known attribute given its visual features, e.g. $p(``mob"|x)$; we then use
  the trained context model to estimate the conditional probability $P(``violence"|``mob")$ between novel and known attributes.}\label{fig:Scheme}
\end{figure}

More precisely, in this chapter we develop a zero-shot multi-label
attribute contextual prediction model (Fig.~\ref{fig:Scheme}). We make
the assumption that the detection of known attributes helps the
recognition of unknown ones. For instance, a putative unknown
attribute such as ``\textit{violence}'' may be related to known
attributes ``\textit{outdoor}'', ``\textit{fight}'', ``\textit{mob}'', and
``\textit{police}'' among others. Therefore, high confidence in these
attributes would support the existence of
``\textit{violence}''. Specifically, our model first learns a
probabilistic $P$-way classifier on $\const{P}$ known attributes,
e.g. $p(``outdoor"|\vect{x})$. Then we estimate the probability of each
novel (unseen) attribute conditioned on the confidence of $P$ known
attributes, e.g. $p(``violence"|``outdoor")$. 
Recall that due to ``\textit{violence}'' in this example being a novel
attribute, this conditional probability cannot be estimated directly
by tabulation of annotation statistics. To model this conditional, we
consider two contextual learning approaches. The first approach relies
on the semantic relatedness between the two attributes. For instance, if ``\textit{fight}'' is semantically related
to ``\textit{violence}'', then we would assume a high conditional
probability $p(``violence"|``fight")$. Crucially, such semantic
relations can be learned in the absence of annotated video
data. This is achieved by using large text corpora
\cite{GanLYZH_AAAI_15} and language models
\cite{Mikolov2013a,pennington2014glove}. However, this text-only based
approach has the limitation that linguistic relatedness may not
correspond reliably to the visual contextual co-occurence that we wish
to exploit. For example, the word ``\textit{outdoor}'' has high
linguistic semantic relatedness, e.g. measured by a cosine similarity,
with ``\textit{indoor}'', whilst they would never co-occur in video
annotations. Therefore, our second approach to conditional probability
estimation is based on learning to map from {\em pairwise} linguistic semantic
relatedness to visual co-occurence. Specifically, on the known
training attributes, we train a bilinear mapping to map {\em the pair of}
training word-vectors (e.g. $\vect{v}(``fight")$ and $\vect{v}(``mob")$) to the training attributes' co-occurrence. This
bilinear mapping can then be used to better predict the conditional
probability between known and novel/unseen attributes. This is
  analogous to the standard ZSL idea of learning a visual-semantic
  mapping from a set of single attributes and re-using this mapping across different
  unseen attributes. Here, we focus instead on a set of
  attribute-pairs to learn co-occurrence mapping, and re-using this
  pairwise mapping across new attribute pairs.

As a proof-of-concept case study, we
consider the task of violent behaviour (event) detection in
videos. This task has received increasing interest in recent years 
\cite{hassner2012violent}, but it is challenging due to the difficulty
of obtaining violent event videos for training reliable
recognisers. In this chapter, we demonstrate our approach by training our
model on an independent large {\em WWW} crowd-attribute video dataset,
which does not contain ``\textit{violence}'' as a known attribute, and
then apply the model to violent event detection on the {\em Violent
  Flow} video dataset \cite{hassner2012violent}. 

In summary, we make the following contributions in this chapter: (1) For
the first time, we investigate zero-shot learning for crowd behaviour
recognition to overcome the costly and semantically ambiguous video
annotation of multi-labels. (2) We propose a contextual learning
strategy which enhances novel attribute recognition through
context prediction by estimating attribute-context co-occurrence with
a bilinear model. (3) A proof-of-concept case study is presented to
demonstrate the viability of transferring zero-shot recognition of
violent event cross-domain with very promising performance.

\section{Related Work}

\subsection{Crowd Analysis}
Crowd analysis is one of the central topics in computer vision
research for surveillance \cite{gong2011security}. There are a variety
of tasks including: (1) Crowd density estimation and person counting
\cite{Chen_featuremining,LoyEtAl_Chapter2013}, (2) crowd tracking
\cite{ali2008floor,rodriguez2011data}, and (3) crowd behaviour
recognition
\cite{andrade2006modelling,wang2009unsupervised,Shao2015}. There are
several challenges in crowd behaviour analysis. First of all, one
requires both informative and robust visual features from crowd
videos. Although simple optical flow
\cite{Xu:2013:CTS:2510650.2510657,rodriguez2011data,saleemi2010scene},
tracklets \cite{zhao2012tracking,zhou2012coherent}, or a combination
of motion and static features \cite{li2012learning} have been
adopted. None of them is both informative and robust. More desirable Scene-level
features can be further constructed from these low-level features,
using probabilistic topic models
\cite{wang2009unsupervised,Xu:2013:CTS:2510650.2510657} or Gaussian
mixtures \cite{saleemi2010scene}. However, these mid-level
representations are mostly scene-specific, with a few exceptions such
as \cite{Xu:2013:CTS:2510650.2510657} which models multiple scenes to
learn a scene-independent representation. Second, for recognition in
different scenes, existing methods rely heavily upon the assumption of the
availability of sufficient observations (a large number of
repetions with variations) from these scenes in order to either
learn behaviour models from scratch
\cite{wang2009unsupervised,li2012learning,saleemi2010scene}, or
inherit models from related scenes
\cite{Xu:2013:CTS:2510650.2510657}. To generalize models across
scenes, studies have proposed scene-invariant crowd/group descriptors
inspired by socio-psychological and biological research
\cite{shao2014scene}, and more recently from deep learning mined
crowd features \cite{Shao2015}. In addition to these purpose-built
crowd features, dense trajectory features \cite{Wang2014} capturing
both dynamic (motion boundary) and static textural information have
also been adopted for crowd analysis \cite{Shao2015}. For learning a
scene-invariant model, the method of \cite{Shao2015} requires
extensive manual annotation of crowd attributes: The {\em WWW} crowd
video dataset \cite{Shao2015} has 94 attributes captured by over
10,000 annotated crowd videos, where each crowd video is annotated
with multiple attributes. The effort required for annotating these
videos is huge. This poses significant challenge to scale up
the annotation of any larger video dataset from diverse
domains. Third, often the most interesting crowd behaviour is also
novel in a given scene/domain. That is, the particular behavioural attribute
has not been seen previously in that domain. To address these
challenges, in this study we explore a different approach to crowd behaviour
recognition, by which crowd attribute context is learned from
a large body of text descriptions rather than relying on exhaustive
visual annotations, and this semantic contextual knowledge is
exploited for zero-shot recognition of novel crowd behavioural
attributes without labelled training samples.
 

\subsection{Zero-Shot Learning}
Zero-shot learning (ZSL) addresses the problem of constructing
recognizers for novel categories without labelled training data
(unseen) \cite{Lampert2009}. ZSL is made possible by leveraging an
intermediate semantic space that bridges visual features and class
labels (semantics). In general, the class labels can be obtained by
manually labelled attributes \cite{Lampert2009,Fu2015}, word-vector
embeddings \cite{Socher2013,Xu_SES_ICIP15}, structured word databases
such as the WordNet \cite{Rohrbach2011,GanLYZH_AAAI_15}, and co-occurrence
statistics from external sources \cite{Mensink2014}.  

\vspace{0.1cm}\noindent\textbf{Attributes}\quad Attributes are
manually defined binary labels of mid-level concepts
\cite{Lampert2009} which can be used to define high-level classes, and
thus bridge known and unknown classes. Traditional supervised
classifiers can be trained to predict attributes rather than
categories. In the testing phase, recognisers for new classes can then
be defined based on novel classes' attributes, e.g. Direct Attribute
Prediction (DAP) \cite{Lampert2009}, or relations to known classes by
the attributes, e.g. Indirect Attribute Prediciton (IAP)
\cite{Lampert2009}. This intuitive attribute based strategy inspired
extensive research into ZSL. However, attributes themselves are
manually annotated and thus suffer from: (i) The difficulty of
determining an appropriate ontology of attributes; (ii) Prohibitive
annotation cost, in particular for videos due to their spatio-temporal
nature; and (iii) labelling each video with a large vocabulary of
attributes is particularly costly and ambiguous. 

Note that attributes in the context of a ZSL \emph{semantic
  representation} are different from the attributes we aim to predict
in this chapter. In the attribute-ZSL case, all attributes are
pre-defined and annotated in order to train supervised classifiers to
generate a representation that bridges known and un-known high-level
classes for \emph{multi-class} ZSL prediction. In our case, we want to
predict multiple  crowd attributes for each video. That is, our final
goal is \emph{multi-label} ZSL prediction, as some of these attributes
are zero-shot, i.e. not pre-defined or annotated in training data.

\vspace{0.1cm}\noindent\textbf{WordNet}\quad 
As an alternative to attributes, WordNet \cite{fellbaum1998wordnet} is
a large English lexical database which organises words in groups (aka
synsets). WordNet is notably exploited for the graph structure which
provides a direct relatedness measurement between classes as a path
length between concepts \cite{Rohrbach2010,GanLYZH_AAAI_15}. The IAP
model can be implemented without attribute annotation by replacing the
novel to known class relation by WordNet induced relation. However,
due to the absence of explicit representation for each individual
word, the WordNet semantics are less likely to generalize to ZSL
models with alternative training losses (e.g. ranking loss and
regression loss) which require explicit embedding of words. 

\vspace{0.1cm}\noindent\textbf{Co-occurrence}\quad 
Studies have also explored external sources for measuring the relation
between known and novel classes. In particular, web hit count has been
considered as a source to induce a co-occurrence based representation
\cite{Rohrbach2010,Mensink2014}. Intuitively, two labels/concepts are
treated closely related if they often co-occur in search engine
results. As with the WordNet based approaches, co-occurrence models
are not able to produce explicit representations for classes therefore
are not compatible with learning alternative losses.

\vspace{0.1cm}\noindent\textbf{Word-Vector}\quad 
The word-vector representation \cite{Mikolov2013a,Socher2013} generated by unsupervised learning on text corpora 
has emerged as a promising representation for ZSL in that: (i) As the
product of unsupervised learning on existing text corpora, it avoids
manual annotation bottlenecks; (ii) Semantic similarity between
words/phrases can be measured as cosine distance in the word-vector
space thus enables probabilistic views of zero-shot learning,
e.g. DAP\cite{Lampert2009} and semantic inter-relations
\cite{GanLYZH_AAAI_15}, and training with alternative models,
e.g. ranking loss \cite{akata2015outputEmbedding,frome2013devise} and
regression loss \cite{Socher2013,Xu_SES_ICIP15}.

\subsection{Multi-Label Learning}

Due to the multiple aspects of crowd behaviour to be detected/recognised, videos
are often annotated with more than one attribute. The multi-attribute
nature of crowd video, makes crowd behaviour understanding a {\em
  multi-label learning} (MLL) \cite{zhang2014review} problem. MLL
\cite{zhang2014review} is the task of assigning a single instance
simultaneously to multiple categories. MLL can be decomposed into a
set of independent single-label problems to avoid the complication of
label correlation \cite{zhang2007ml,boutell2004learning}. Although
this is computationally efficient, ignoring label correlation produces
sub-optimal recognition. Directly tackling the joint multi-label
problem through considering all possible label combinations is
intractable, as the size of the output space and the required training
data grow exponentially w.r.t. the number of unique labels \cite{tsoumakas2007random}. 
As a compromise, tractable solutions to correlated multi-label
prediction typically involve considering {\em pairwise} label
correlations
\cite{furnkranz2008multilabel,qi2007correlative,ueda2002parametric},
e.g. using conditional random fields (CRF)s. 
However, all existing methods require to learn these pairwise label
correlations in advance from the statistics of large labeled
datasets. In this chapter, we solve the challenge of multi-label
prediction for labels without any existing annotated datasets from
which to extract co-occurrence statistics.

\subsection{Multi-Label Zero-Shot Learning}
Although zero-shot learning is now quite a well studied topic, only a
few studies have considered multi-label zero-shot learning
\cite{Fu2014,Mensink2014}. Joint multi-label prediction is challenging
because conventional multi-label models require pre-computing the
label co-occurrence statistics, which is not available in the ZSL
setting. The study given by \cite{Fu2014} proposed a Direct
Multi-label zero-shot Prediction (DMP) model. This method synthesises
a power-set of potential testing label vectors so that visual features
projected into this space can be matched against every possible
combination of testing labels with simple NN matching. This is
analogous to directly considering the jointly multi-label problem,
which is intractable due to the size of the label power-set growing
exponentially ($2^n$) with the number of labels being considered. 
An alternative study was provided by \cite{Mensink2014}. Although
applicable to the multi-label setting, this method used co-occurrence
statistics as the semantic bridge between visual features and class
names, rather than jointly predicting multiple-labels that can
disambiguate each other. A related problem is to jointly predict
multiple attributes when attributes are used as the semantic embedding
for ZSL \cite{hariharan2012efficient}. In this case, the
correlations of mid-level attributes, which are multi-labelled, are
exploited in order to improve single-label ZSL, rather than the inter-class
correlation being exploited to improve multi-label ZSL. 


\section{Methodology}

We introduce in this section a method for recognising novel crowd
behavioural attributes by exploring the context from other
recognisable (known) attributes. In section~\ref{sect:PGM}, we
introduce a general procedure for predicting novel behavioural
attributes based on their relation to known attributes. This is
formulated as a probabilistic graphic model adapted from
\cite{Lampert2009} and \cite{gan2015devnet}. We then give the details in
section~\ref{sect:SuppLabelCorrection} on how to learn a behaviour predictor that
estimates the relations between known and novel attributes by inferring
from text corpus and co-occurrence statistics of known attribute annotations. 


Let us first give an overview of the notations used in this
chapter in Table~\ref{tab:Notations}. Formally we have training dataset
$\mathcal{T}^{S}=\{\matr{X}^{S},\matr{Y}^{S},\matr{V}^{S}\}$ associated with $P$
known attributes and testing dataset
$\mathcal{T}^{T}=\{\matr{X}^{T},\matr{Y}^{T},\matr{V}^{T}\}$ associated with $Q$
novel/unseen attributes. We denote the visual feature for training and
testing videos as $\matr{X}^{S}=[\vect{x}_1,\cdots
  \vect{x}_\const{N_S}]\in \mathbb{R}^{\const{D}_x \times
  \const{N_S}}$ and $\matr{X}^{T}=[\vect{x}_1,\cdots
  \vect{x}_\const{N_T}]\in \mathbb{R}^{\const{D}_x \times
  \const{N_T}}$, multiple binary labels for training and testing
videos as $\matr{Y}^{S}=[\vect{\tilde{y}}_1,\cdots
  \vect{\tilde{y}}_\const{N_S}]\in \{0,1\}^{\const{P}\times
  \const{N_S}}$ and $\matr{Y}^{T}=[\vect{y^*}_1,\cdots
  \vect{y^*}_\const{N_T}]\in \{0,1\}^{\const{Q}\times
  \const{N_T}}$, and the continuous semantic embedding (word-vector)
for training and testing attributes as $\matr{V}^{S}=[\vect{v}_1
  \cdots \vect{v}_\const{P}]\in \mathbb{R}^{\const{D_v}\times
  \const{P}}$ and $\matr{V}^{T}=[\vect{v}_1 \cdots
  \vect{v}_\const{Q}]\in \mathbb{R}^{\const{D_v}\times
  \const{Q}}$. Note that according to the zero-shot assumption, the
training and testing attributes are disjoint i.e. $\forall p\in
\{1\cdots P\},q\in \{1\cdots Q\}:\vect{v}_p\in \matr{V}^S,
\vect{v}_q\in \matr{V}^T, \vect{v}_p\neq
\vect{v}_q$. 

\begin{table}[t!]
\centering
\caption{Notation Summary}
\label{tab:Notations}
\resizebox{0.95\linewidth}{!}{ 
\revise{
\begin{tabular}{ll}
\hline
\textbf{Notation}                   & \textbf{Description}                     \\ \hline
$N_S; ~N_T $					&	Number of training/source instances ; testing/target instances	\\
$D_x; ~D_v$						& Dimension of visual feature; of word-vector embedding	\\
$P; ~Q$						& Number of training/source attributes ; testing/target attributes\\
$\matr{X}\in \mathbb{R}^{D_x\times N}$; ~$\vect{x}$     & Visual feature matrix for N instances; column representing one instance   \\
$\matr{Y}\in \{0,1\}^{P\times N}$; ~$\vect{y}$      & Binary labels for $N$ instances with $P$ (or $Q$) labels; column representing one instance      \\
$\matr{V}\in \mathbb{R}^{D_v\times P}$; ~$\vect{v}$     & Word-Vector embedding for $P$ (or $Q$) attributes; column representing embedding for one attribute    \\
 \hline
\end{tabular}}}
\end{table}

\subsection{Probabilistic Zero-Shot Prediction}\label{sect:PGM}
\begin{figure}
\centering
\includegraphics[width=0.6\linewidth]{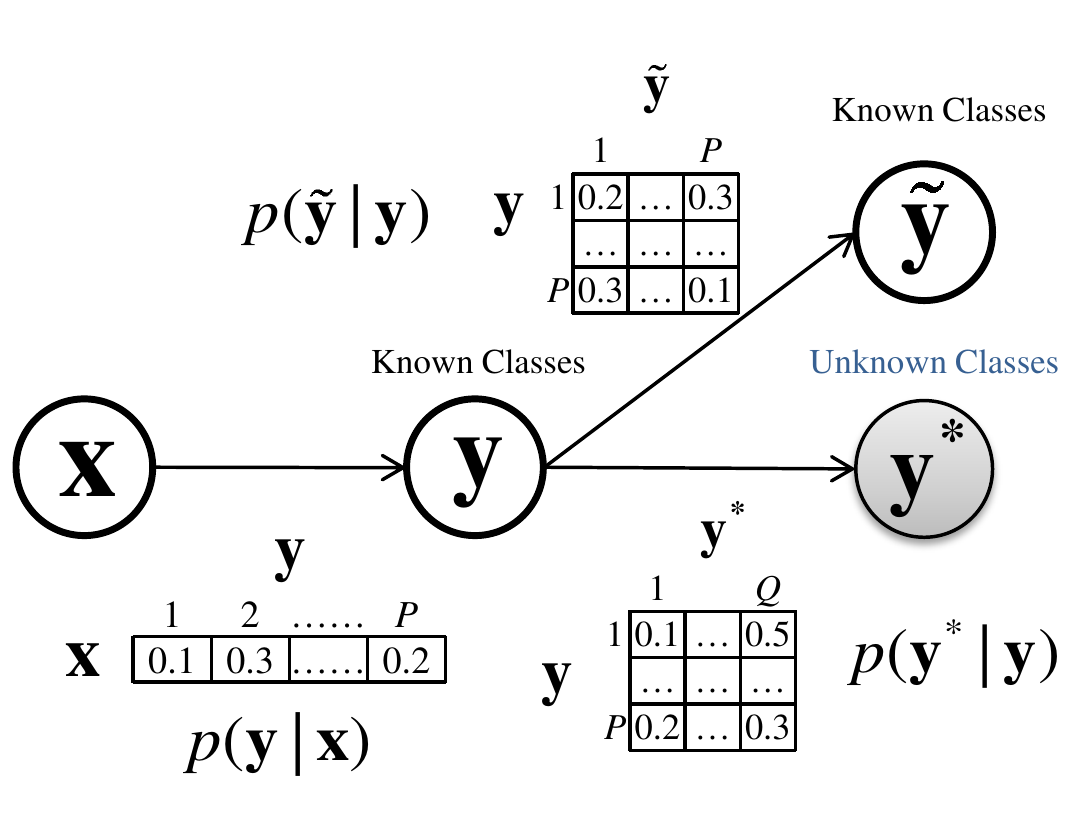}
\caption{A probablistic graphical representation of a context-aware
  multi-label zero-shot prediction model.} 
\label{fig:PGM}
\end{figure}

To predict novel attributes by reasoning about the relations between
known and novel attributes, we formulate this reasoning process
as a probabilistic graph (see Fig.~\ref{fig:PGM}).

Given any testing video $\vect{x}$, we wish to assign it with one or
many of the $\const{P}$ known attributes or $\const{Q}$ novel
attributes. This problem is equivalent to inferring a set of
conditional probabilities
$p(\vect{y^*}|\vect{x})=\{p({y}^*_q|\vect{x})\}_{q=1\cdots \const{Q}}$
and/or
$p(\vect{\tilde{y}}|\vect{x})=\{p(\tilde{y}_p|\vect{x})\}_{p=1\cdots
  \const{P}}$. To achieve this, given the video instance $\vect{x}$,
we first infer the likelihood of it being one of the $\const{P}$ known
attributes as $p(\vect{y}|\vect{x})=\{p(y_p|\vect{x})\}_{p=1\cdots
  P}$. Then, given the relation between known and novel/known
attributes as conditional probability $P(\vect{y}^*|\vect{y})$ or
$P(\vect{\tilde{y}}|\vect{y})$, we formulate the conditional
probability similar to Indirect Attribute Prediction (IAP)
\cite{Lampert2009,GanLYZH_AAAI_15} as follows:

\begin{equation}\label{eq:MarginProb}
\begin{split}
p({y}^*_q|\vect{x})=\sum\limits_{p=1}^{\const{P}}p({y}^*_q|y_p)p(y_p|\vect{x})\\
p(\tilde{y}_{\tilde{p}}|\vect{x})=\sum\limits_{p=1}^{\const{P}}p(\tilde{y}_{\tilde{p}}|y_p)p(y_p|\vect{x})
\end{split}
\end{equation}
The zero-shot learning task is to infer the probabilities
$\{p({y}^*_q|\vect{x})\}_{p=1\cdots \const{P}}$ for unseen labels
$\{y^*_q\}$. We estimate the multinomial conditional probability of
known attributes $p(y_p|\vect{x})$ based on the output of a
probabilistic P-way classifier, e.g. SVM or Softmax Regression with
probability output. Then the key to the success of zero-shot
prediction is to estimate the known to novel contextual attribute
relation as conditional probabilities $\{p({y}^*_q|y_p)\}$. We
introduce two approaches to estimate this contextual relation.

\subsection{Modelling Attribute Relation from Context}\label{sect:SuppLabelCorrection}

In essence, our approach to the prediction of novel attributes depends
on the prediction of known attributes and then predicting the novel
attributes based on the confidence of each known attribute. The key to
the success of this zero-shot prediction is therefore appropriately
estimating the conditional probability of novel attribute given known
attributes. We first consider a more straightforward way to model this
conditional by exploiting the relation encoded by a {\em text} corpus
\cite{GanLYZH_AAAI_15}. We then extend this idea to predict the
expected \emph{visual} co-occurrence between novel and known
attributes without labelled samples of the novel classes.

\subsubsection{Learning Attribute Relatedness from Text Corpora}\label{sect:TexCAZSL}

The first approach builds on semantic word embedding
\cite{GanLYZH_AAAI_15}. The semantic embedding represents each English
word as a continuous vector $\vect{v}$ by training a skip-gram neural
network on a large text corpus \cite{Mikolov2013a}. The objective of
this neural network is to predict the \revise{adjacent $c$ words to the
current word $w_t$}, as follows:

\begin{equation}\label{eq:WordVectorObj}
\frac{1}{\const{T}}\sum\limits_{t=1}^\const{T}\sum\limits_{-c\leq j\leq c,j\neq 0} p(w_{t+j}|w_t)
\end{equation}

\noindent The conditional probability is modelled by a softmax
function, a normalised probability distribution, based on each word's representation as a continuous vector:

\begin{equation}
p(w_{t+j}|w_t)=\frac{exp(\vect{v}_{t+j}^\top \vect{v}_t)}{\sum\limits_{j=1}^{W}exp(\vect{v}_{t+j}^\top \vect{v}_t)}\label{eq:predSoftmax}
\end{equation} 
By maximizing the above objective function, the learned word-vectors
$\matr{V}=\{\mathbf{v}\}$ capture contextual co-occurrence in the
text corpora so that frequently co-occurring words result in high
cumulative log probability in Eq~(\ref{eq:WordVectorObj}). We apply
the softmax function to model conditional attribute probability \revise{ as Eq~(\ref{eq:WVLikelihood}) where $\gamma$ is a temperature parameter}.

\begin{equation}\label{eq:WVLikelihood}
p({y}^*_q|y_p)=\frac{exp(\frac{1}{\gamma}\vect{v}_q^\top \vect{v}_p^S)}{\sum\limits_{p=1}^{P}exp(\frac{1}{\gamma}\vect{v}_q^\top \vect{v}^S_p)}
\end{equation} 
This can be understood intuitively from the
following example: An attribute ``\textit{Shopping}'' has high affinity with
attribute ``\textit{ShoppingMall}'' in word-vector inner product
because they co-occur in the text corpus. Our assumption is that the
existence of known video attribute ``\textit{Shopping}'' would support
the prediction of unseen attribute ``\textit{ShoppingMall}''. 

\subsubsection{Context Learning from Visual Co-Occurrence}\label{sect:CoCAZSL}

Although  attribute relations can be discovered from text context as
described above, these relations may {\em not} ideally suit crowd
attribute prediction in videos. For example, the inner product of
$vec(``Indoor")$ and $vec(``Outdoor")$ is $0.7104$ which is ranked the
1st w.r.t. ``Indoor'' among 93 attributes in the {\em WWW} crowd video
dataset. As a result, the estimated conditional probability
$p(\tilde{y}_{Indoor}|y_{Outdoor})$ is the highest among all
$\{p(\tilde{y}_{Indoor}|y_p)\}_{p=1\cdots\const{P}}$. However, whilst
these two attributes are similar because they occur nearby in the text
semantical context, it is counter-intuitive for
visual co-occurrence as a video is very unlikely to be {\em both} indoor and
outdoor. Therefore in visual context, their conditional probability should be small rather than large.  

To address this problem, instead of directly paramaterising the
conditional probability using word-vectors, we use pairs of word vectors to
\emph{predict} the actual visual co-occurrence. More precisely, we train a
word-vector$\to$co-occurrence predictor based on an auxiliary set of
known attributes annotated on videos, for which both word-vectors and
annotations are known. We then re-deploy this learned predictor for
zero-shot recognition on novel attributes.  
Formally, given binary multi-label annotations $\matr{Y}^S$ on
training video data, we define the contextual attribute occurrence as
$\matr{C}={\matr{Y}^S}{\matr{Y}}^{S\top}$. The occurrence of $j$-th attribute in the context of $i$-th attribute is thus $\scal{c}_{ij}$ of the $\matr{C}$. The
prevalence of $i$-th attribute is defined as 
$\scal{c}_i=\sum_j \scal{c}_{ij}$. The normalized co-occurrence thus defines the
conditional probability as:

\begin{equation}
p(\tilde{y}_j|\tilde{y}_i)=\frac{c_{ij}}{c_i}
\end{equation}
The conditional probability can only be estimated based on visual
co-occurrence in the case of training attributes with annotations
$\mathbf{Y}^S$.  To estimate the conditional probability for testing
data of novel attributes without annotations $\mathbf{Y}^T$, we
consider to \emph{predict} the expected co-occurrence based on a
bilinear mapping $\matr{M}$ from the pair of word-vectors.
Specifically, we approximate the un-normalized co-occurrence as
$exp(\vect{v}_i^\top \matr{M} \vect{v}_j)=\scal{c}_{ij}$. To estimate
$\matr{M}$, we optimise the regularised linear regression problem:

\begin{equation}
J = \sum\limits_{i}^P\sum\limits_{j}^P
w(\scal{c}_{ij})\left(\vect{v}_i^\top \matr{M} \vect{v}_j - \log
\scal{c}_{ij}\right)^2 + \lambda ||\matr{M}||_F^2,\label{eq:objective}
\end{equation}

\noindent \revise{where $\lambda$ is the regularisation strength, and} a weight function $w(c_{ij})$ is applied to the regression loss
function above in order to penalise rarely occurring co-occurrence
statistics. We choose the weight function according to
\cite{Mensink2014}, which is:
\begin{equation}\label{eq:WeightFcn}
w(c_{ij}) = \left(\frac{c_{ij}}{\const{C_{max}}}\right)^{(\alpha \cdot \mathbbm{1}(c_{ij}\leq \const{C}_{max}))}
\end{equation}
where $\const{C}_{max}$ is a threshold of co-occurrence, \revise{$\alpha$ controls the increasing rate of the weight function} and the  $\mathbbm{1}$ is an indicator function.
This bilinear mapping  is related to the model in
\cite{pennington2014glove}, but differs in that: (i) The input of the
mapping is the word-vector representations $\vect{v}$ learned from the
skip-gram  model \cite{Mikolov2013a} in order to generalise to novel
attributes where no co-occurrence statistics are available. (ii) The
mapping is trained to account for  \emph{visual} compatibility,
e.g. ``{\em Outdoor}'' is unlikely to co-occur with ``{\em Indoor}''
in a visual context, although the terms are closely related in their
representations learned from the text corpora alone. The bilinear
mapping can be seamlessly integrated with the softmax conditional
probability as:

\begin{equation}\label{eq:CooccurrenceSoftmax}
p(y^*_q|y_p)=\frac{exp(\vect{v}_q^\top \matr{M} \vect{v}_p)}{\sum\limits_{p} exp(\vect{v}_q^\top \matr{M} \vect{v}_p)}
\end{equation}
Note that by setting $\matr{M}=\matr{I}$, this conditional
probability degenerates to the conventional word-vector based estimation in
Eq~(\ref{eq:WVLikelihood}). 
The regression to predict visual co-occurrence from word-vectors
(Eq.~(\ref{eq:objective})) can be efficiently solved by gradient
descent using the following gradient: 

\begin{equation}
\nabla \vect{M} =
\sum\limits_{i=1}^{\const{P}}\sum\limits_{j=1}^{\const{P}}f(\scal{c}_{ij})\left(2\vect{v}_i\vect{v}_i^\top
\matr{M} \vect{v}_j\vect{v}_j^\top -
2\log\scal{c}_{ij}\vect{v}_i\vect{v}_j^\top\right) + 2\lambda \matr{M} 
\end{equation}

\section{Experiments}
We evaluate our multi-label crowd behaviour recognition model on the
large {\em WWW} crowd video dataset \cite{Shao2015}. We analyse each
component's contribution to the overall multi-label ZSL
performance. Moreover, we present a proof-of-concept case study for
performing transfer zero-shot recognition of violent behaviour in the
{\em Violence Flow} video dataset \cite{hassner2012violent}.

\subsection{Zero-Shot Multi-Label Behaviour Inference}

\subsubsection{Experimental Settings}\label{sect:ExpSetting}

\begin{figure}[]
\centering
\subfloat[27 attributes by ``Where'']{\includegraphics[width=0.9\linewidth]{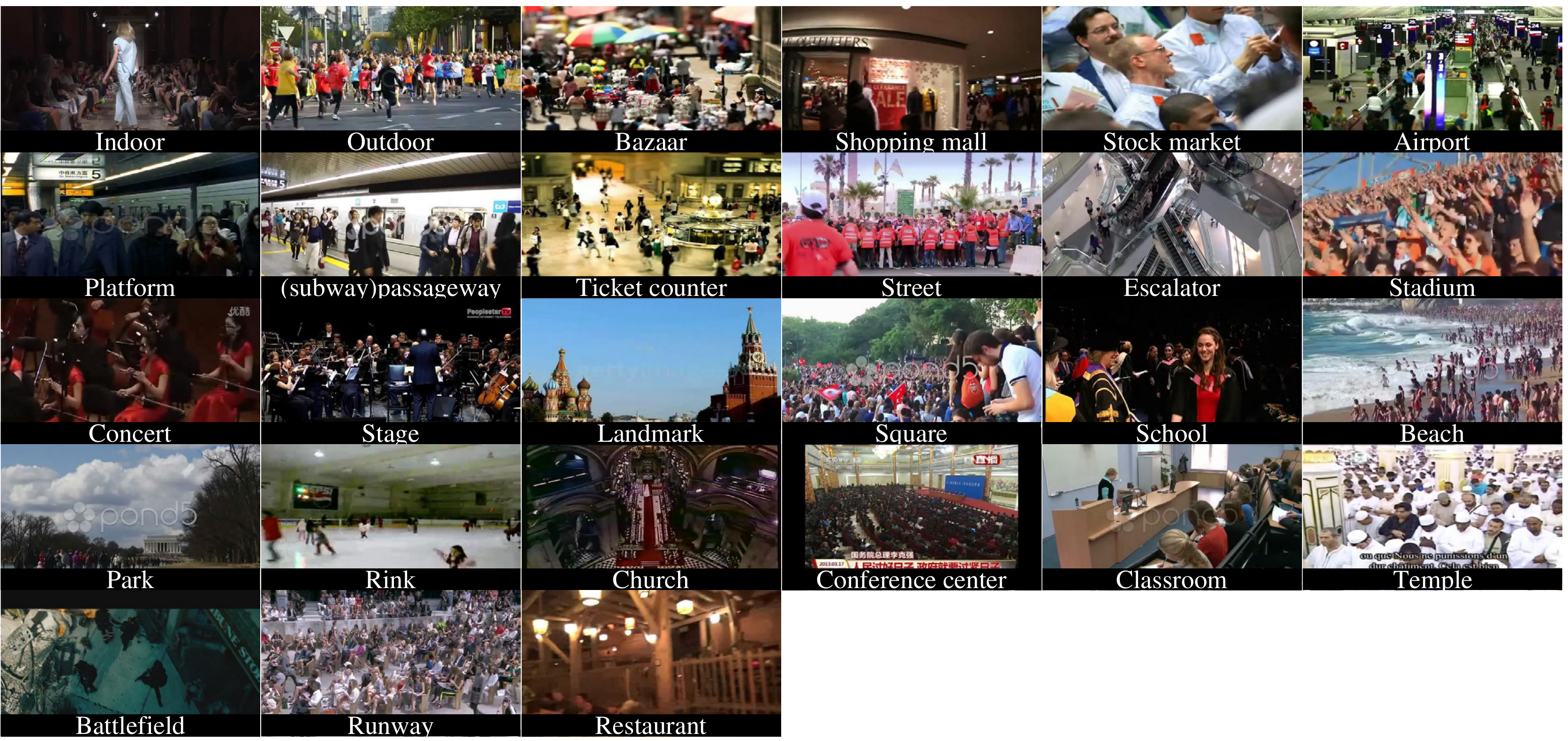}}\\
\subfloat[24 attributes by ``Who'']{\includegraphics[width=0.9\linewidth]{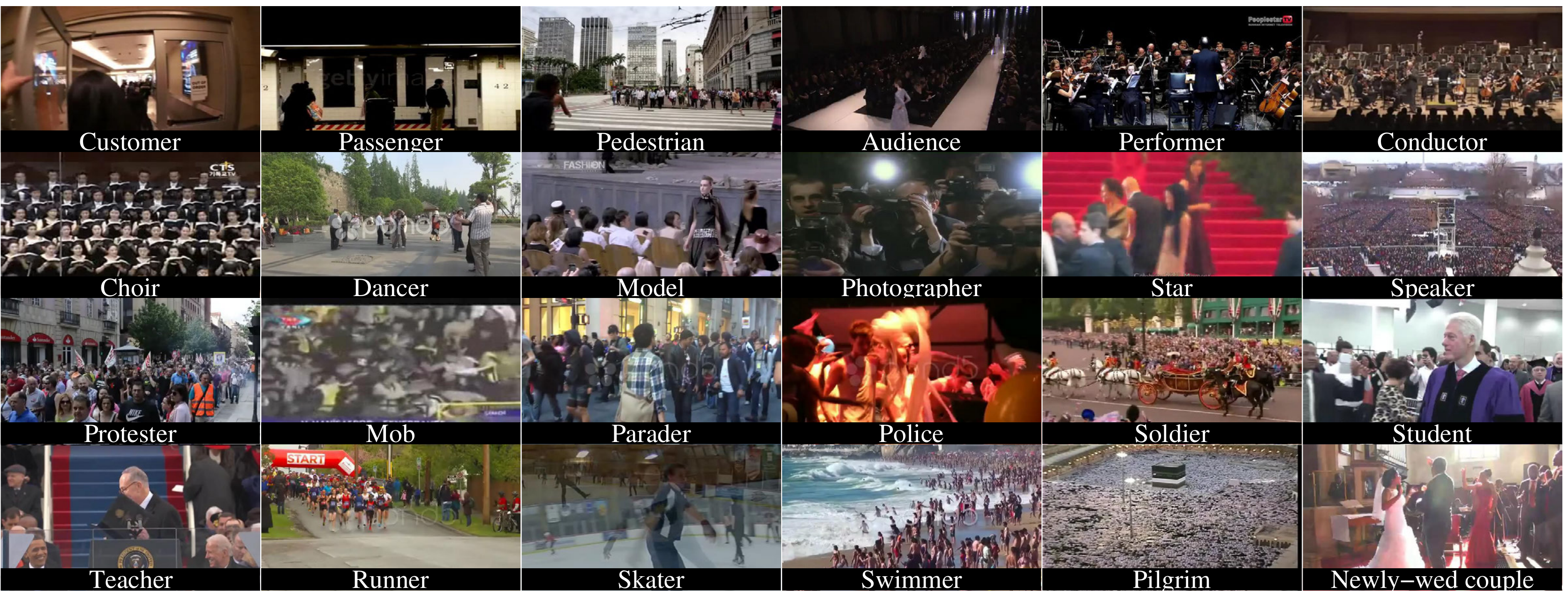}}\\
\subfloat[44 attributes by ``Why'']{\includegraphics[width=0.9\linewidth]{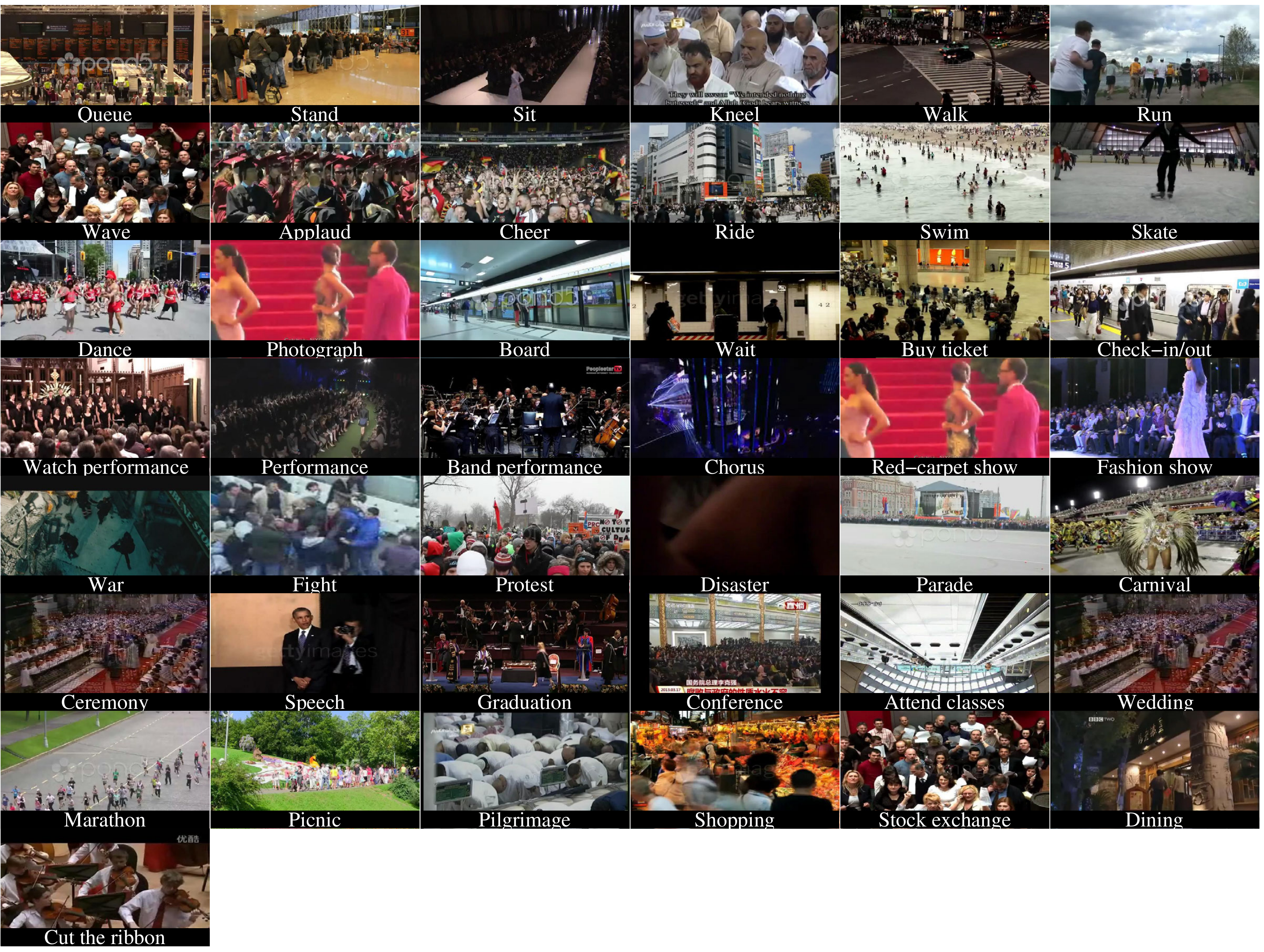}}
\caption{Examples of all attributes in the {\em WWW}
  crowd video dataset \cite{Shao2015}.}\label{fig:WWW_example_frame}
\end{figure}

\paragraph{\textbf{Dataset}}
The {\em WWW} crowd video dataset is specifically proposed for
studying scene-independent attribute prediction for crowd scene
analysis. It consists of over 10,000 videos collected from online resources
of 8,257 unique scenes.  The crowd attributes are designed to answer
the following questions: ``Where is the crowd'', ``Who is in the crowd'' and
``Why is the crowd here''.  All videos are manually annotated with 94
attributes with 6 positive attributes per video on
average. Fig.\ref{fig:WWW_example_frame} shows a collection of 94
examples with each example illustrating each attribute in the {\em WWW} crowd video dataset.

\begin{figure}
\centering
\subfloat[Label distribution over all videos]{\includegraphics[width=0.48\linewidth]{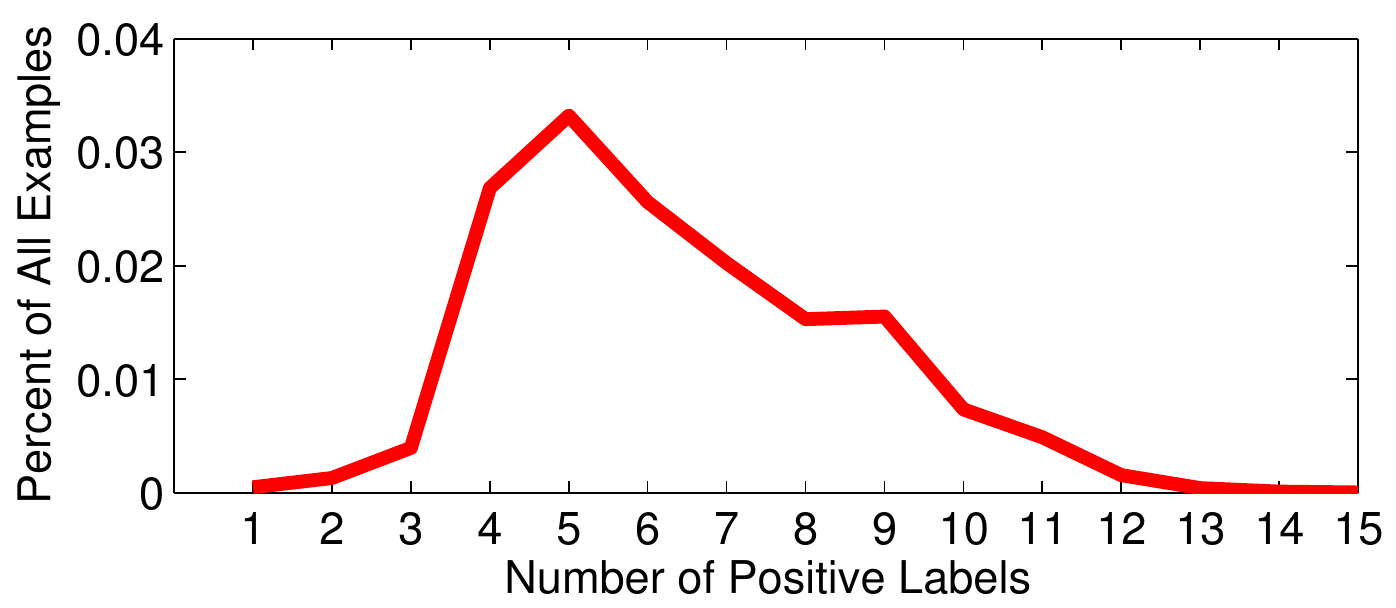}}
\subfloat[Label distribution over testing videos]{\includegraphics[width=0.48\linewidth]{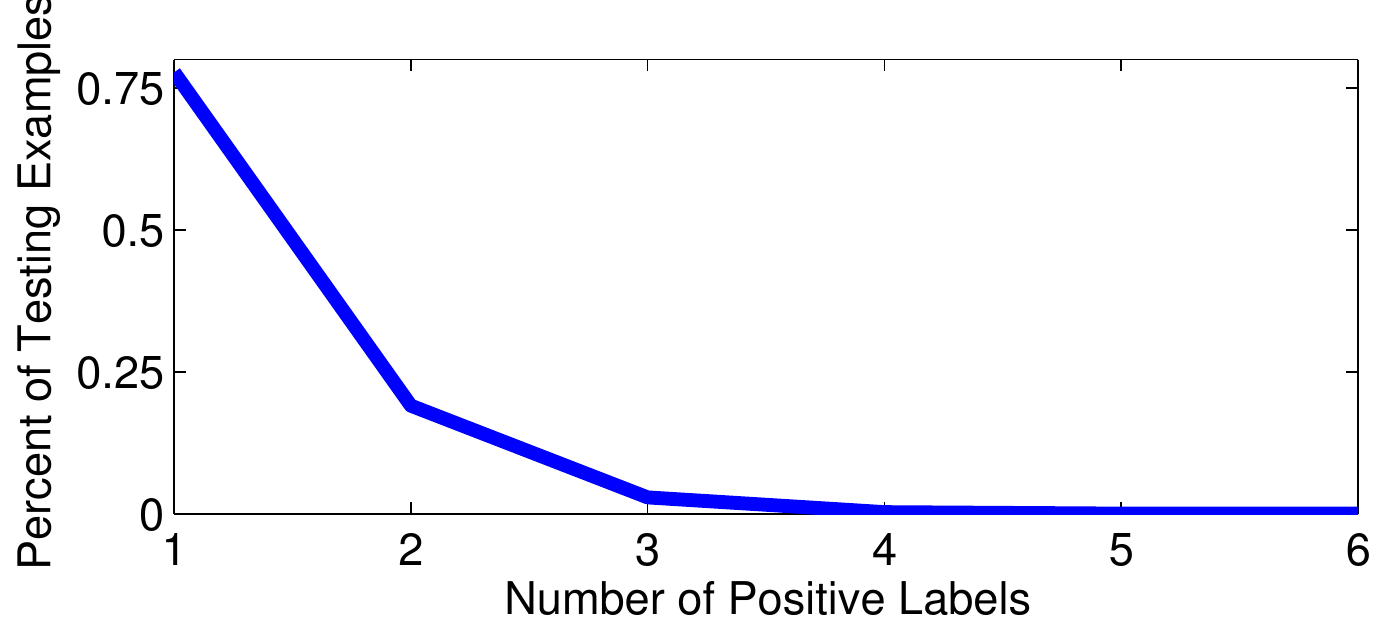}}\\
\subfloat[Distribution of testing videos]{\includegraphics[width = 0.55\linewidth]{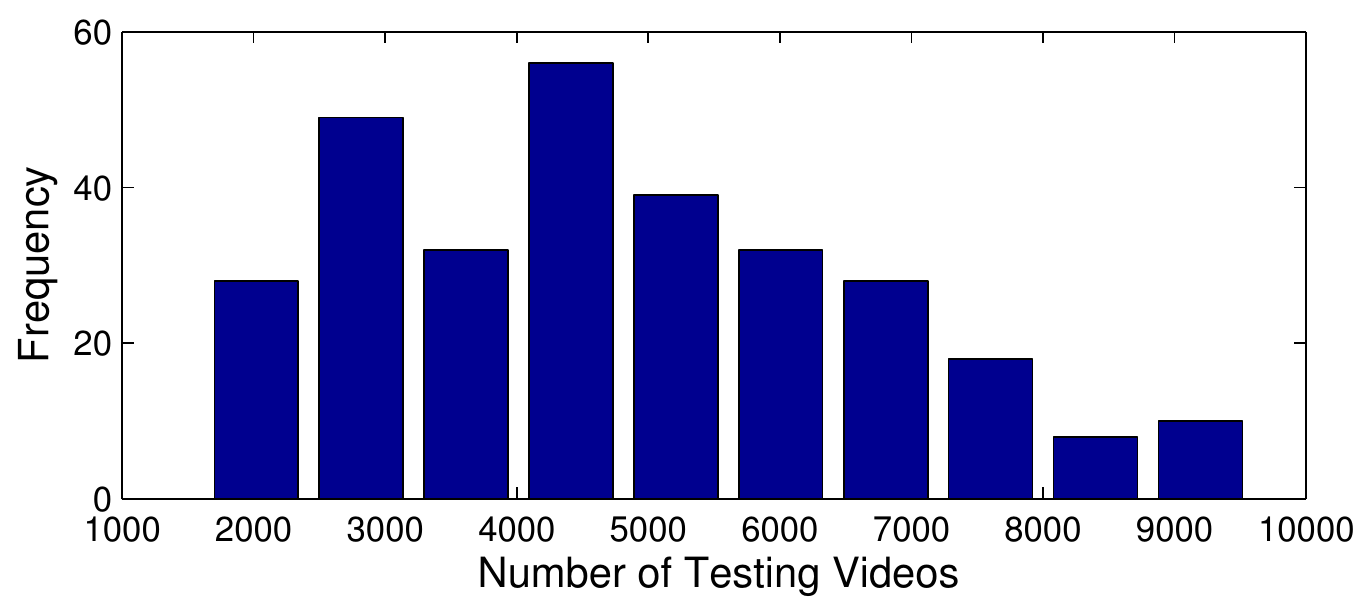}}
\caption{Statistics of the dataset split for our experiments on the
  {\em WWW} dataset. (a) and (b): The distributions of
  multi-label per video over all the videos and over the testing
  videos respectively. (c): The distribution of the number of
  testing videos over all 50 random splits.}\label{fig:DistAttr} 
\end{figure}

\paragraph{\textbf{Data Split}}
We validated the ability to utilise known attributes for recognising
novel attributes in the absence of training samples on the {\em WWW}
dataset. To that end, we divided the 94 attributes into 85 for
training (known) and 9 for testing (novel). This was repeated for 50
random splits. In every split, any video which has no positive label
from the 9 novel attributes was used for training and the rest for
testing. The distributions of the number of multi-attributes (labels) per video over all
videos and over the testing videos are shown in
Fig~\ref{fig:DistAttr}(a-b) respectively. Fig~\ref{fig:DistAttr}(c)
also shows the distribution of the number of testing
videos over the 50 random splits. In most splits, the number of testing
videos is in the range of 3,000 to 6,000. The training to testing video
number ratio is between 2:1 to 1:1. This low training-testing ratio makes for a
challenging zero-shot prediction setting. 

\paragraph{\textbf{Visual Features}}
Motion information can play an important role in crowd scene
analysis. To capture crowd dynamics, we extracted the improved dense
trajectory features \cite{Wang2014} and performed Fisher vector
encoding \cite{Perronnin2010} on these features, generating a $50,688$
dimensional feature vector to represent each video. 

\paragraph{\textbf{Evaluation Metrics} }

We evaluated the performance of multi-label prediction using five
different metrics \cite{zhang2014review}. These multi-label prediction
metrics fall into two groups: Example-based metric and label-based
metric. Example-based metrics evaluate the performance per video instance
and then average over all instances to give the final
metric. Label-based metrics evaluate the performance per label category and
return the average over all label categories as the final metric. The
five multi-label prediction performance metrics are:

\begin{itemize}
\item \textbf{AUC} - The Area Under the ROC Curve. AUC evaluates
  binary classification performance. It is invariant to the
  positive/negative ratio of each testing label. Random guess
  leads to AUC of 0.5. For multi-label prediction, we measure the AUC for
  each testing label and average the AUC over all 50 splits to yield 
  the AUC per category. The final mean AUC is reported as the mean
  over all lable categories. 

\item \textbf{Label-based AP} - Label-based Average Precision. We
  measure the average precision for each attribute as the average
  fraction of relevant videos ranked higher than a
  threshold. The random guess baseline for label-based AP is
  determined by the prevalence of positive videos. 

\item \textbf{Example-based AP} - Example-based Average Precision. We
  measure the average precision for each video as the average fraction
  of relevant label prediction ranked higher than a
  threshold. Example-based AP focuses on the rank of attributes 
  within each instance rather than rank of examples for each label as
  for label-based AP. 

\item \textbf{Hamming Loss} - Hamming Loss measures the percentage of
  incorrect predictions from groundtruth labels. Optimal hamming
  loss is 0, indicating perfect prediction. Due to the
  nature of hamming loss, the distance of $[0 0 0]$ and $[1 1 0]$
  w.r.t. $[0 1 0]$ are equal. Thus it does not differentiate
  over-estimation from under-estimation. Hamming loss is a label-based 
  metric. The final mean is reported as the average over all
  instances. 

\item \textbf{Ranking Loss} - Ranking Loss measures, for every
  instance, the percentage of negative labels ranked higher than
  positive labels among all possible positive-negative label
  pairs. Similar to example-based AP, the ranking loss is
  example-based metric focusing on pushing positive labels ahead
  of negative labels for each instance. 

\end{itemize}

Both AUC and label-based AP are label-based metrics, whilst
exampled-based AP, Hamming Loss and Ranking Loss are example-based
metrics. Moreover, as a loss metric, both Hamming Loss and Ranking Loss
values are lower the better. In contrary, AUC and AP values are higher the better.  
In a typical surveillance application of crowd behaviour recognition
in videos, we are interested in detecting video instances of
a particular attribute that triggers an alarm event, e.g. searching
for video instances with the ``\textit{fighting}'' attribute. In this
context, label-based performance metrics such as AUC and Label-based
AP are more relevant. Overall, we present model performance evaluated
by both types of evaluation metrics.

\revise{
\paragraph{\textbf{Parameter Selection}}
We have several parameters to tune in our model. Specifically, for training SVM classifiers for known classes/attributes $\{p(y|\vect{x})\}$ we set the slack parameter to a constant 1. The ridge regression coefficient $\lambda$ in Eq~(\ref{eq:objective}) is essential to avoid over-fitting and numerical problems. It is empirically set to small non-zero value. We choose $\lambda=1^{-3}$ in our experiments. For the temperature parameter $\gamma$ in Eq~(\ref{eq:WVLikelihood}), we cross validate and found best value to be around $0.1$. In addition, we use a word-vector dictionary pre-trained on Google News dataset \cite{Mikolov2013a} with 100 billion words where word-vectors are trained with 300 dimension ($D_v=300$) and context size 5 ($c=5$).
}


\subsubsection{Comparative Evaluation}\label{sect:WWW_ZSL}

In this first experiment, we evaluated zero-shot multi-label
prediction on {\em WWW} crowd video dataset. We compared our
context-aware multi-label ZSL models, both purely text-based and
visual co-occurrence based, against four contemporary and
state-of-the-art zero-shot learning models.

\paragraph{\textbf{Sate-of-the-art ZSL Models}}
\begin{enumerate}
\item Word-Vector Embedding (\textbf{WVE}) \cite{Xu_SES_ICIP15}: The
  WVE model constructs a vector representation
  $\vect{z}_{tr}=g(\scal{y}_{tr})$ for each training instance
  according to its category name $\scal{y}_{tr}$ via word-vector
  embedding $g(\cdot)$ and then learns a support vector regression
  $f(\cdot)$ to map the visual feature $\vect{x}_{tr}$. For testing
  instance $\vect{x}_{te}$, it is first mapped into the semantic
  embedding space via the regressor $f(\vect{x}_{te})$. Novel category
  $\scal{y}_{te}\in \mathcal{Y}_{te}=\{1,\cdots Q\}$ is then mapped
  into the embedding space via $g(\scal{y}_{te})$. Nearest neighbour
  matching is applied to match $\vec{x}_{te}$ with category
  $\scal{y}^*$ using the L2 distance:

\begin{equation}
\scal{y}^*=\arg\min\limits_{y_{te}\in \mathcal{Y}_{te}} ||f(\vect{x}_{te})-g(\scal{y}_{te})||_2^2
\end{equation}
We do not assume having access to the whole testing data distribution,
so we do not exploit transductive self-training and data augmentation
post processing, unlike in the cases of \cite{Xu_SES_ICIP15,AlexiouXG_ICIP16}.

\item Embarrassingly Simple Zero-Shot Learning (\textbf{ESZSL})
  \cite{romera2015embarrassingly}: The ESZSL model considers  ZSL as
  training a L2 loss classifier. Specifically, given known categories'
  binary labels $\matr{Y}$ and word-vector embedding $\matr{V}_{tr}$,
  we minimise the L2 classification loss as: 

\begin{equation}
\min\limits_\matr{M} \sum\limits_{i=1}^{\const{N}}||\vect{x}_i^\top \matr{M} \matr{V}_{tr} - \vect{y}_i||^2_2 + \Omega(\matr{M};\matr{V}_{tr},\matr{X})
\end{equation}

\revise{where $\Omega(\matr{M};\matr{V}_{tr},\matr{X})$ is a regulariser defined as:}

\revise{
\begin{equation}
\Omega(\matr{M};\matr{V}_{tr},\matr{X})=\lambda_1||\matr{M}\matr{V}_{tr}||_F^2+\lambda_2||\matr{X}^{\top}\matr{M}||_F^2+\lambda_3||\matr{M}||_F^2
\end{equation}
}

 Novel categories are predicted by:

\begin{equation}\label{eq:ESZSL_Predict}
\vect{y}^* = \vect{x}_{te}^\top \matr{M} \vect{V}_{te}
\end{equation}

\item Extended DAP (\textbf{ExDAP}) \cite{Fu2014}: ExDAP was
  specifically proposed for multi-label zero-shot learning
  \cite{Fu2014}. This is an extension of single-label
  regression models to multi-label. Specifically, given training
  instances $\vect{x}_i$, associated multiple binary labels
  $\vect{y}_i$, and word-vector embedding of known labels
  $\matr{V}_{tr}$, we minimize the L2 regression loss for learning a
  regressor $\matr{M}$: 

\begin{equation}\label{eq:TrExDAP}
\min\limits_{\matr{M}}\sum\limits_{i=1}^\const{N}||\vect{x}_i^\top \matr{M} - \matr{V}_{tr}\vect{y}_i||_2^2+\lambda ||\matr{M}||_2^2
\end{equation}

For zero-shot prediction, we minimize the same loss but w.r.t. the binary label vector $\vect{y}$ with L2 regularization:

\begin{equation}
\vect{y}^*=\arg\min\limits_{\vect{y^*\in \mathbb{R}}}||\vect{x}_{te}^\top \matr{M} - \matr{V}_{te}\vect{y^*}||_2^2+\lambda ||\vect{y}^*||_2^2
\end{equation}

A closed-form solution exists for prediction:

\begin{equation}\label{eq:ExDAP_Predict}
\vect{y}^*=\left(\matr{V}_{te}^\top\matr{V}_{te}+\lambda\matr{I}\right)^{-1}\matr{V}_{te}^\top\vect{x}_{te}^\top \matr{M}
\end{equation}

\item Direct Multi-Label Prediction (\textbf{DMP}) \cite{Fu2014}: DMP
  was proposed to exploit the correlation between testing labels so to
  benefit the multi-label prediction. It shares the same training
  procedure with ExDAP in Eq~(\ref{eq:TrExDAP}). For zero-shot
  prediction, given testing categories $\mathcal{Y}_{te}$ we first
  synthesize a power-set of all labels
  $\mathcal{P}(\mathcal{Y}_{te})$. The multi-label prediction
  $\vect{y}^*$ is then determined by  nearest neighbour matching of
  visual instances mapped into word-vector embedding
  $\vect{x}_{te}^\top\matr{M}$ against the  synthesized power-set: 

\begin{equation}
\vect{y}^*=\arg\min\limits_{\vect{y}^*\in \mathcal{P}(\mathcal{Y}_{te})} ||\vect{x}_{te}^\top\matr{M} - \matr{V}_{te}\vect{y}^*||_2^2
\end{equation}

\end{enumerate}


\paragraph{\textbf{Context-Aware Multi-Label ZSL Models}}
\begin{enumerate}
\item Text Context-Aware ZSL (\textbf{TexCAZSL}): In our text corpus
  context-aware model introduced in Section~\ref{sect:TexCAZSL}, only
  word-vectors learned from text corpora \cite{Mikolov2013a} are used
  to model the relation between known and novel attributes
  {$p(\vect{y}^*|\vect{y})$}, as defined by Eq~(\ref{eq:WVLikelihood}). We implemented the video instance to known attributes
  probabilities {$p({y}_p|\vect{x})$} as $\const{P}$ linear SVM
  classifiers with normalized probability outputs \cite{CC01a}. Novel
  attribute prediction $p({y}^*_q|\vect{x})$ is computed by
  marginalising over the known attributes defined by
  Eq~(\ref{eq:MarginProb}). 

\item Visual Co-occurrence Context-Aware ZSL (\textbf{CoCAZSL}): We
  further implemented a visual co-occurrence context-aware model built
  on top of the \textbf{TexCAZSL} model. This is done by predicting
  the expected co-occurrence context using bilinear mapping
  $\matr{M}$, as introduced in Section~\ref{sect:CoCAZSL}. The known
  to novel attribute relation is thus modelled by a weighted
  inner-product between the word-vectors of known and novel attributes
  given by Eq~(\ref{eq:CooccurrenceSoftmax}). Novel attribute
  prediction $p({y}^*_q|\vect{x})$ is computed in the same way as
  for \textbf{TexCAZSL}, defined by Eq~(\ref{eq:MarginProb}).

\end{enumerate}

%
%
%

\paragraph{\textbf{Quantitative Comparison}}

Table~\ref{tab:ComparisonZSLWWW} shows the comparative results of our
models against four state-of-the-art ZSL models and the baseline of
``Random Guess'', using all five evaluation metrics. Three
observations can be made from these results: (1) All zero-shot
learning models can substantially outperform random guessing,
suggesting that zero-shot crowd attribute prediction is valid. This
should inspire more research into zero-shot crowd behaviour analysis
in the future. (2) It is evident that our context-aware models
improve on existing ZSL methods when measured by the label-based AUC
and AP metrics. As discussed early under evaluation
metrics, for typical surveillance tasks, label-based metrics provide a
good measurement on detecting novel alarm events in the mist of many
other contextual attributes in crowd scenes. (3) It is also evident
that our context-aware models perform comparably to the alternative
ZSL models under the example-based evaluation metrics, with the
exception that DMP \cite{Fu2014} performs extraordinarily well on
Hamming Loss but poorly on Ranking Loss. This is due to the direct
minimization of Hamming Loss between synthesized power-set and
embedded video in DMP. However, since the relative order
between attributes are ignored in DMP, low performance in ranking loss
as well as other label-based metrics is expected. 

\begin{table}[h!]
\centering
\caption{Comparison of zero-shot multi-label attribute prediction on
  the {\em WWW} crowd video dataset. The $\uparrow$ and $\downarrow$
  symbols indicate whether a metric is higher the better or vice versa.}
\label{tab:ComparisonZSLWWW}
\begin{tabular}{l|l|c|c|c|c|c}
\toprule
\multicolumn{1}{c|}{\textbf{Feature}} & \multicolumn{1}{c|}{\textbf{Model}}             & \multicolumn{2}{c|}{\textbf{Label-Based}}                   & \multicolumn{3}{c}{\textbf{Example-Based}}                                                             \\ \cline{3-7} 
                             &                                        & \multicolumn{1}{c|}{\textbf{AUC $\uparrow$}} & \multicolumn{1}{c|}{\textbf{AP $\uparrow$}} & \multicolumn{1}{c|}{\textbf{AP $\uparrow$}} & \multicolumn{1}{c|}{\textbf{Hamming Loss $\downarrow$}} & \multicolumn{1}{c}{\textbf{Ranking Loss $\downarrow$}} \\ \hline
-                            & Random Guess                           & 0.50                     & 0.14                    & 0.31                    & 0.50                              & -                                \\ \hline
ITF                          & WVE\cite{Xu_SES_ICIP15}            & 0.65                        & 0.24                       & 0.52                       & 0.45                                 & 0.32                                \\
ITF                          & ESZSL\cite{romera2015embarrassingly} & 0.63                     & 0.22                    & 0.53                       & 0.46                              & 0.32                                \\
ITF                          & ExDAP\cite{Fu2014}                   & 0.62                     & 0.21                    & 0.52                    & 0.45                              & 0.32                             \\
ITF                          & DMP\cite{Fu2014}                     & 0.59                     & 0.20                    & 0.45                    & \textbf{0.30}                              & {0.70}                             \\ \hline
ITF                          & TexCAZSL                               & 0.65                     & 0.24                       & {0.52}                    & 0.43                              & 0.32                             \\
ITF                          & CoCAZSL                                & \textbf{0.69}                     & \textbf{0.27}                       & \textbf{0.53}                    & 0.42                              & \textbf{0.31}                             \\ \bottomrule
\end{tabular}
\end{table}

\paragraph{\textbf{Qualitative Analysis}}

\begin{figure}[!ht]
\centering
\includegraphics[width=0.9\linewidth]{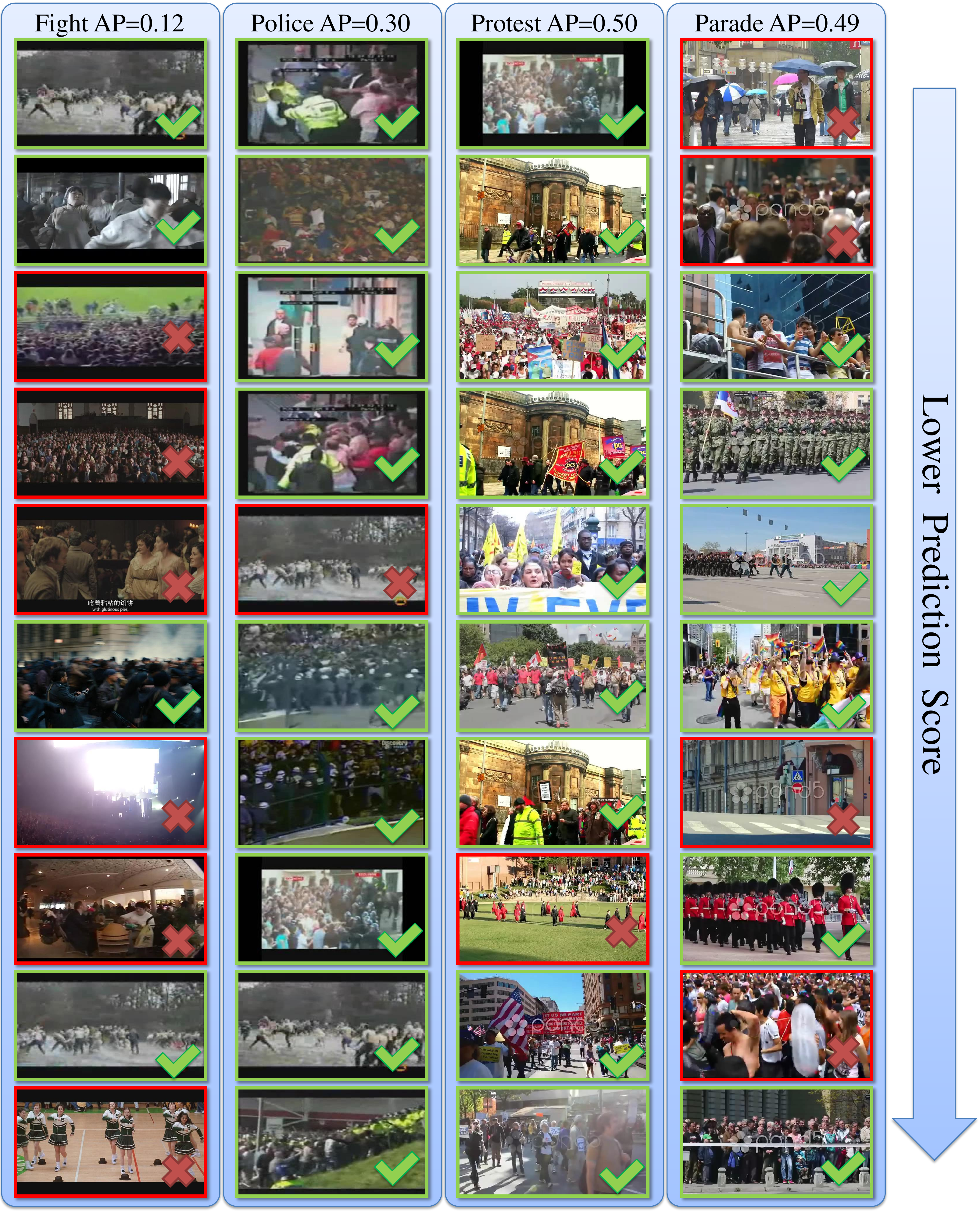}
\caption{Illustration of crowd videos ranked in accordance with prediction scores (marginalized conditional probability) w.r.t. each attribute.}\label{fig:Qualitative_LabelBased}
\end{figure}

We next give some qualitative examples of zero-shot attribute predictions in Fig.~\ref{fig:Qualitative_LabelBased}. To get a sense of how well the attributes are detected in the context of label-based AP, we present the AP number with each attribute. Firstly, we give  examples of detecting videos matching some randomly chosen attributes (label-centric evaluation). By designating an attribute to detect, we list the crowd videos sorted in the descending order of probability $p(y^*|\vect{x})$. In general, we observe good performance in ranking crowd videos according to the attribute to be detected. The false detections are attributed to the extremely ambiguous visual cues. E.g. 3rd video in ``\textit{fight}'', 5th video in ``\textit{police}'' and 2nd video in ``\textit{parade}'' are very hard to interpret.

\begin{figure}[!htb]
\centering
\includegraphics[width=0.95\linewidth]{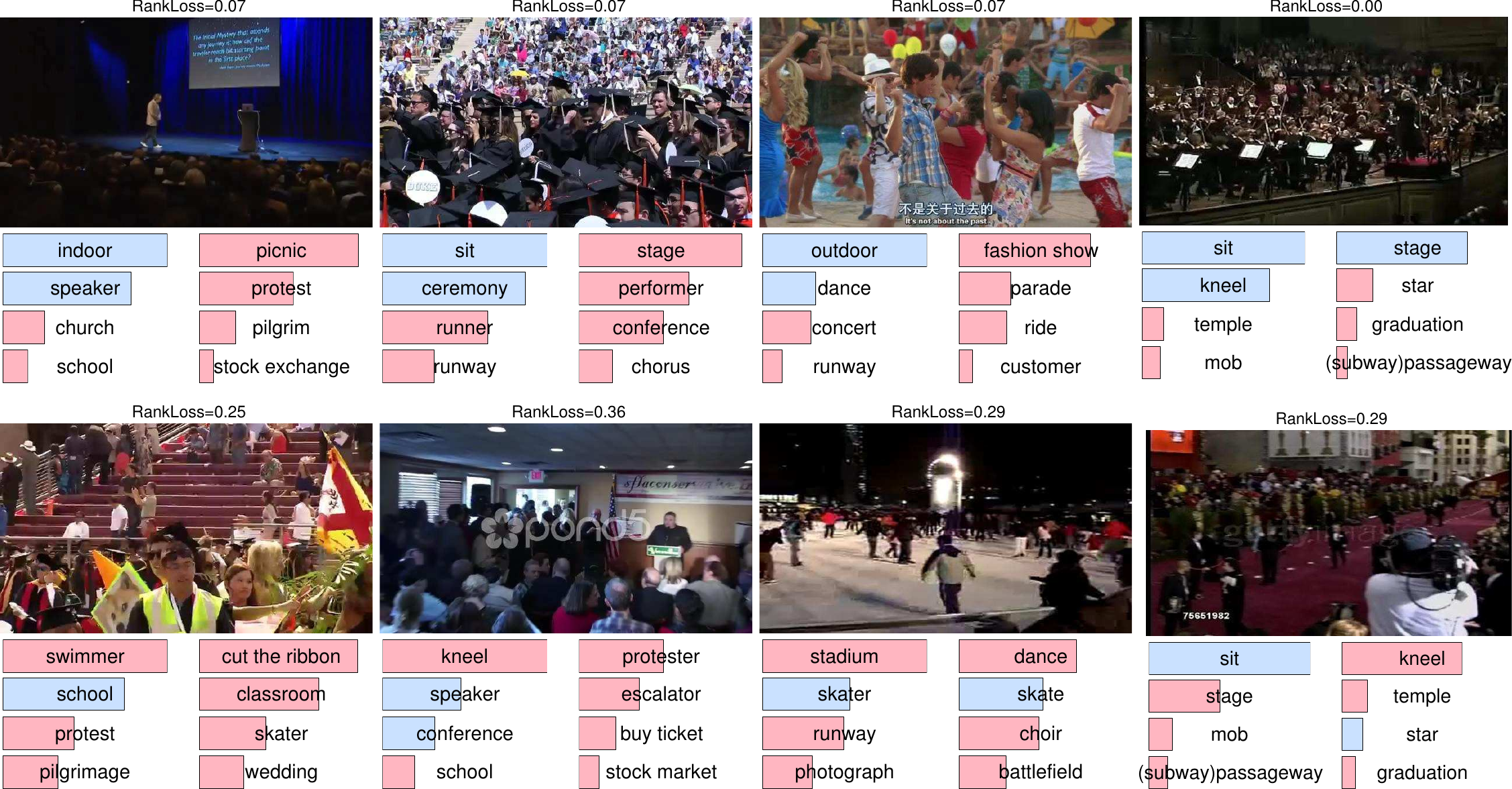}
\caption{Examples of zero-shot multi-label attribute prediction. Bars under each image indicate the normalized score for testing attribtues. Blue and pink bars indicate positive and negative ground-truth labels respectively.}\label{fig:Qualitative_ExampleBased}
\end{figure}

In addition to detecting each individual attribute, we also present some examples of simultaneously predicting multiple attributes in Fig.\ref{fig:Qualitative_ExampleBased} (example-centric evaluation). For each video we give the the prediction score for all testing attributes as $\{p(y^*_q|\vect{x})\}_{q=1\cdots \const{Q}}$ . For the ease of visualization, we omit the the attribute with least score. We present the example-based ranking loss number along with each video  to give a sense of how the quantitative evaluation metric relates to the qualitative results. In general, ranking loss less than $0.1$ would yield very good multi-label prediction as all labels would be placed among the top 3 out of 9 labels to be predicted. Whilst ranking loss around $0.3$ (roughly the average performance of our CoCAZSL model, see Table~\ref{tab:ComparisonZSLWWW}) would still give reasonable predictions by placing positive labels in the top 5 out of 9.

\subsection{Transfer Zero-Shot Recognition in Violence Detection}

Recognizing violence in surveillance scenario has an important role in safety and security \cite{hassner2012violent,gong2011security}. However due to the sparse nature of violent events in day to day surveillance scenes, it is desirable to exploit  zero-shot recognition to detect violent events without human annotated training videos. Therefore we explore a proof of concept case study of transfer zero-shot violence detection. We learn  to recognize labelled attributes in WWW dataset \cite{Shao2015} and then transfer the model to detect violence event in Violence Flow dataset \cite{hassner2012violent}. This is zero-shot because we use no annotated examples of violence to train, and violence does not occur in the label set of WWW. It is contextual because the violence recognition is based on the predicted visual co-occurrence between each known attribute in WWW and the novel violence attribute. E.g., ``\textit{mob}'' and ``\textit{police}'' attributes known from WWW may support the violence attribute in the new dataset. 

\subsubsection{Experiment Settings}

\paragraph{\textbf{Dataset}}

\begin{figure}[!h]
\centering
\subfloat[Violent videos]{\includegraphics[width=0.9\linewidth]{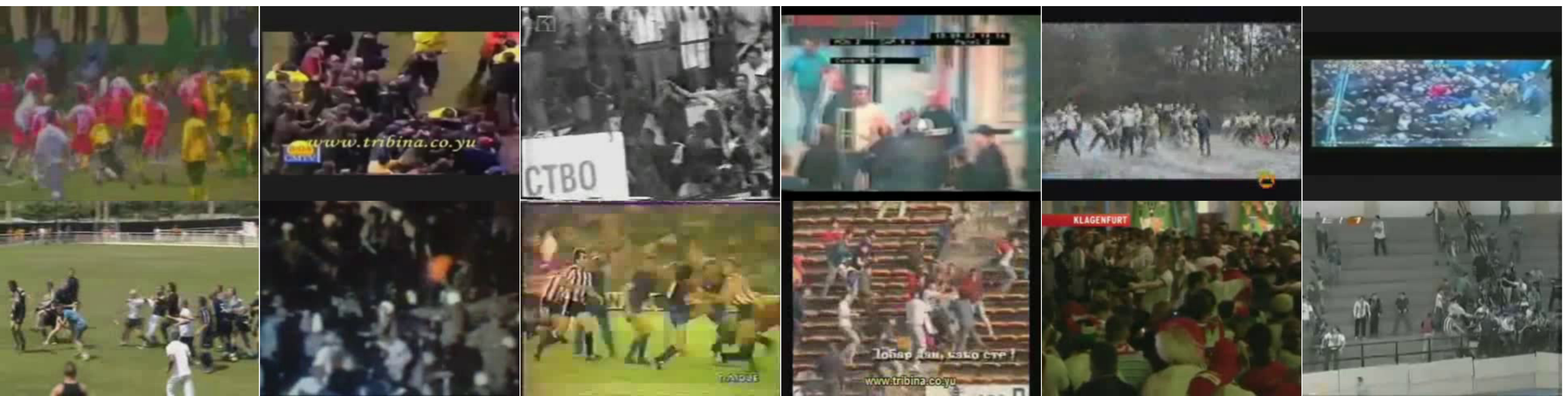}}\\
\subfloat[Non-violent videos]{\includegraphics[width=0.9\linewidth]{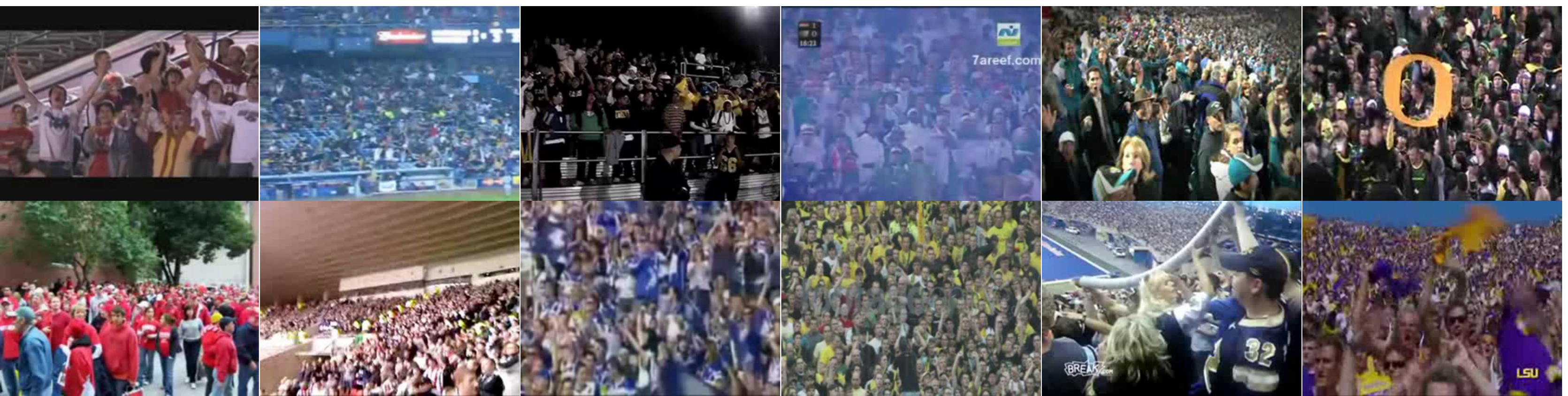}}
\caption{Example frames of violence flow dataset \cite{hassner2012violent}.}\label{fig:ViolenceFlowExample}
\end{figure}

The Violence Flow dataset \cite{hassner2012violent} was proposed to facilitate the study into classifying violent events in crowded scenes. 246 videos in total are collected from online video repositories (e.g. YouTube) with 3.6 seconds length on average. Half of the 246 video are with positive violence content and the another half are with non-violent crowd contents. We illustrate example frames of both violent and non-violent videos in Fig.~\ref{fig:ViolenceFlowExample}

\paragraph{\textbf{Data Split}}
A standard fully supervised 5-fold cross validation split was proposed by \cite{hassner2012violent}. The standard split partitions the whole dataset into 5 splits each of which is evenly divided into positive and negative videos. For each testing split, the other 4 splits are used as the training set and the left-out one is the testing set. Results are reported as both the mean classification accuracy over 5 splits plus standard deviation and the area under the ROC curve (AUC).

Beyond the standard cross validation split we create a new zero-shot experimental design. Our zero-shot split learns attribute detection models on all 94 attributes from WWW dataset and then tests on the same testing set as the standard 5 splits in \cite{hassner2012violent}. We note that there are 123 overlapped videos between WWW and Violence Flow. To make fair comparison, we exclude these overlapped videos from constructing the training data for 94 attributes. In this way zero-shot prediction performance can be directly compared with supervised prediction performance using AUC metric. We define the event/attribute to be detected as the word ``\textit{violence}''. 

\paragraph{\textbf{Zero-Shot Recognition Models}}
We explore the transfer zero-shot violence recognition by comparing the same set of zero-shot learning models as in Section~\ref{sect:WWW_ZSL}: competitors WVE, ESZSL, ExDAP; and our TexCAZSL and CoCAZSL. 

\paragraph{\textbf{Fully Supervised Model}}

To put zero-shot recognition performance in context, we also report fully supervised models' performance. These models are evaluated on the 5-fold cross-validation split and the average accuracy and AUC are reported. Specifically, we report the best performance of \cite{hassner2012violent} - linear SVM with VIolent Flows (ViF) descriptor and our fully supervised baseline - linear SVM with Improved Trajectory Feature (ITF).

\paragraph{\textbf{Results and Analysis}}

The results of both transfer zero-shot and supervised violence prediction are summarised in Table~\ref{tab:Combine_ViolenceFlow}. We make the following observations: Our context-aware models perform consistently better than alternative zero-shot models, suggesting that context does facilitate zero-shot recognition. Surprisingly, our zero-shot models moreover perform very competitively compared to the fully supervised models. Our \textbf{CoCAZSL} (albeit with better ITF feature) beats the fully supervised Linear SVM with ViF feature in AUC metric (87.22 v.s. 85.00). The context-aware model is also close to the fully supervised model with the same ITF feature (87.22 v.s. 98.72).  This is in contrast to the common result in the literature were zero-shot recognition ``works'', but does so much worse than fully supervised learning. The promising performance is partly due to modelling the co-occurrence on large known crowd attributes help the correct prediction of known to novel attribute relation prediction.
Overall the result shows that by transferring our attribute recognition model trained for a wide set of 94 attributes on a large 10,000 video dataset, it is possible to perform effective zero-shot recognition of a novel behaviour type in a new dataset.


\begin{table}[!ht]
\centering
\caption{Evaluation of violence prediction in Violence Flow dataset: zero-shot versus fully supervised prediction ($\%$).}
\label{tab:Combine_ViolenceFlow}
\begin{tabular}{l|c|c|c|c}
\toprule
\multicolumn{1}{c|}{\textbf{Model}} & \textbf{Split}     & \textbf{Feature} & \textbf{Accuracy}       & \textbf{AUC}   \\ \hline
WVE\cite{Xu_SES_ICIP15}                        & Zero-Shot & ITF     & 64.27$\pm$5.06 & 64.25 \\
ESZSL\cite{romera2015embarrassingly}                      & Zero-Shot & ITF     & 61.30$\pm$8.28              & 61.76     \\
ExDAP\cite{Fu2014}                      & Zero-Shot & ITF     & 54.47$\pm$7.37 & 52.31 \\
TexCAZSL                   & Zero-Shot & ITF     & 67.07$\pm$3.87 & 69.95 \\
CoCAZSL                    & Zero-Shot & ITF     & \textbf{80.52$\pm$4.67} & \textbf{87.22} \\ \hline
Linear SVM                 & 5-fold CV & ITF     & 94.72$\pm$4.85 & 98.72 \\
Linear SVM\cite{hariharan2012efficient}                 & 5-fold CV & ViF     & 81.30$\pm$0.21 & 85.00 \\ \bottomrule
\end{tabular}
\end{table}

\section{Further Analysis}

In this section we provide further analysis on the importance of the visual feature used, and also give more insight into how our contextual zero-shot multi-label prediction works by visualising the learned label-relations.

\subsection{Feature Analysis}

We first evaluate different static and motion features on the standard supervised attribute prediction task. Both hand-crafted and deeply learned features are reported for comparison.

\paragraph{\textbf{Static Features}}

We report the both the hand-crafted and deeply learned static feature from \cite{Shao2015} including Static Feature (SFH) and Deeply Learned Static Feature (DLSF). SFH captures general image content by extracting Dense SIFT\cite{Lazebnik2006}, GIST \cite{Oliva2001} and HOG \cite{Dalal2005}. Color histogram in HSV space is further computed to capture global information and LBP \cite{Zhao2012} is extracted to quantify local texture. Bag of words encoding is used to create comparable features, leading to a 1536 dimension static feature. DLSF is initialized using a pre-trained model for ImageNet detection task \cite{ouyang2014deepid} and then fine-tuned on the WWW attribute recognition task with cross-entropy loss.

\paragraph{\textbf{Motion Features}}

We report both the hand-crafted and deeply learned motion features from \cite{Shao2015} including DenseTrack\cite{wang2011action}, spatio-temporal motion patterns (STMP) \cite{kratz2009anomaly} and Deeply Learned Motion Feature (DLMF) \cite{Shao2015}. Apart from the reported evaluations, we compare them with the improved trajectory feature (ITF) \cite{Wang2014} with fisher vector encoding. Though ITF is constructed in the same way as DenseTrack reported in \cite{Shao2015}, we make a difference in that the visual codebook is trained on a collection of human action datasets (HMDB51\cite{Kuehne2011}, UCF101 \cite{Soomro2012}, Olympic Sports \cite{NieblesCL_eccv10} and CCV\cite{conf/mir/JiangYCEL11}).

\paragraph{\textbf{Analysis}}
Performance on the standard WWW split \cite{Shao2015} for static and motion features is reported in Table~\ref{tab:WWWSupervised}. We can clearly observe that the improved trajectory feature is consistently better than all alternative static and motion features. Surprisingly, ITF is even able to beat deep features (DLSF and DLMF). We attribute this to ITF's ability to capture both motion information by motion boundary histogram (MBH) and histogram of flow (HoF) descriptors and texture information by Histogram of Gradient (HoG) descriptor. 

More interestingly, we demonstrate that motion feature encoding model (fisher vector) learned from action datasets can benefit the crowd behaviour analysis. Due to the vast availability of action and event datasets and limited crowd behaviour data, a natural extension work is to discover if deep motion model pre-trained on action or event dataset can help crowd analysis. 


\begin{table}[]
\centering
\caption{Comparison between different visual features for attribute prediction.}
\label{tab:WWWSupervised}
\begin{tabular}{l|c}
\toprule
Alternative Features                 & Mean AUC \\ \hline
SFH \cite{Shao2015}                       & 0.81     \\
DLSF \cite{Shao2015}                     & 0.87     \\
DenseTrack \cite{Shao2015}                & 0.63     \\
DLMF \cite{Shao2015}                      & 0.68     \\
SFH+DenseTrack \cite{Shao2015}            & 0.82     \\
DLSF+DLMF \cite{Shao2015}                & 0.88     \\ \hline
Our Features                                 &          \\ \hline
Improved Trajectory Feature (ITF)           & 0.91     \\\bottomrule
\end{tabular}
\end{table}

\subsection{Qualitative Illustration of Contextual Co-occurrence Prediction}
\begin{figure}
\centering
\subfloat[``violence'' as novel event]{\includegraphics[width=0.75\linewidth]{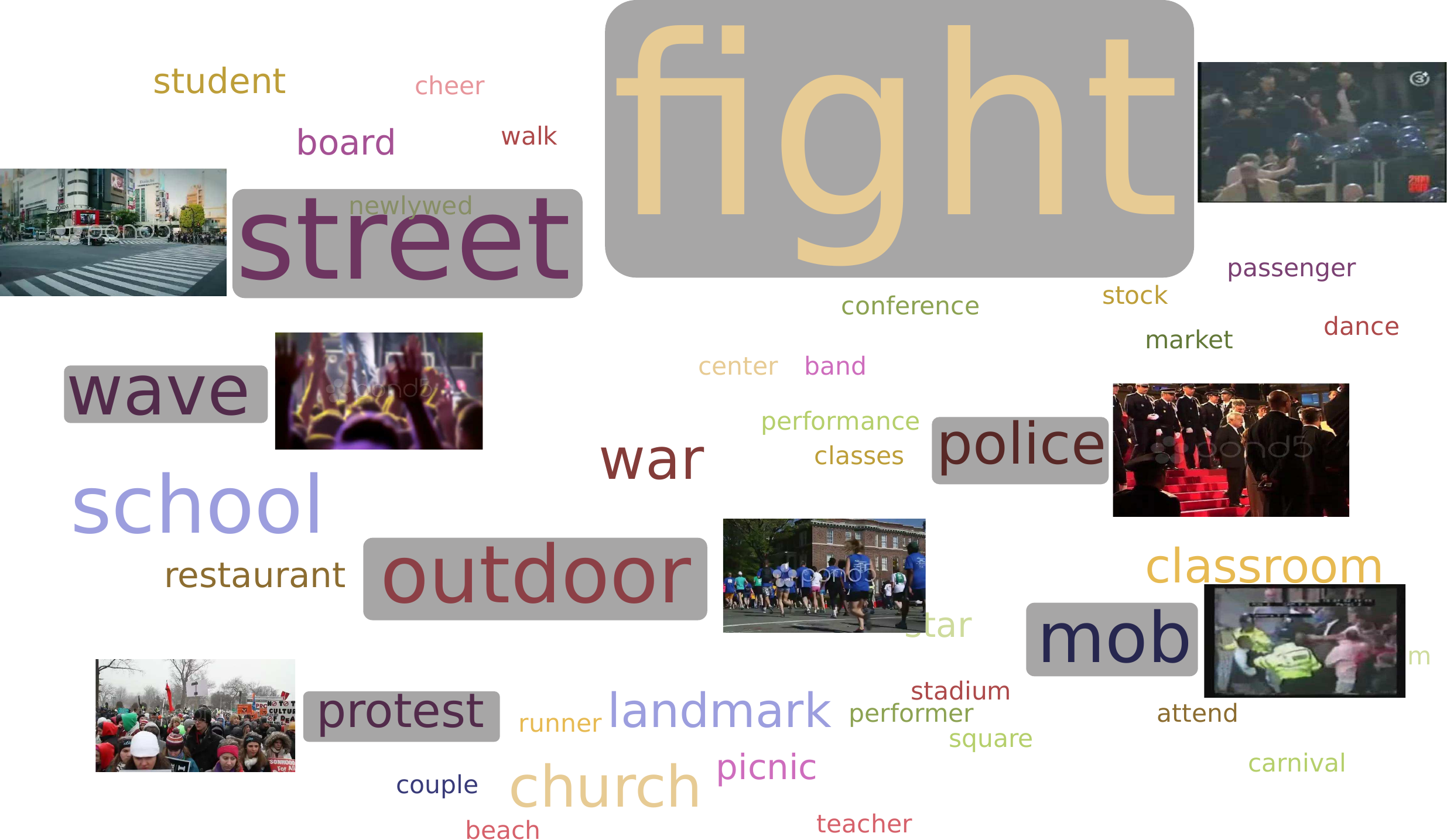}}\\
\subfloat[``mob'' as novel attribute]{\includegraphics[width=0.75\linewidth]{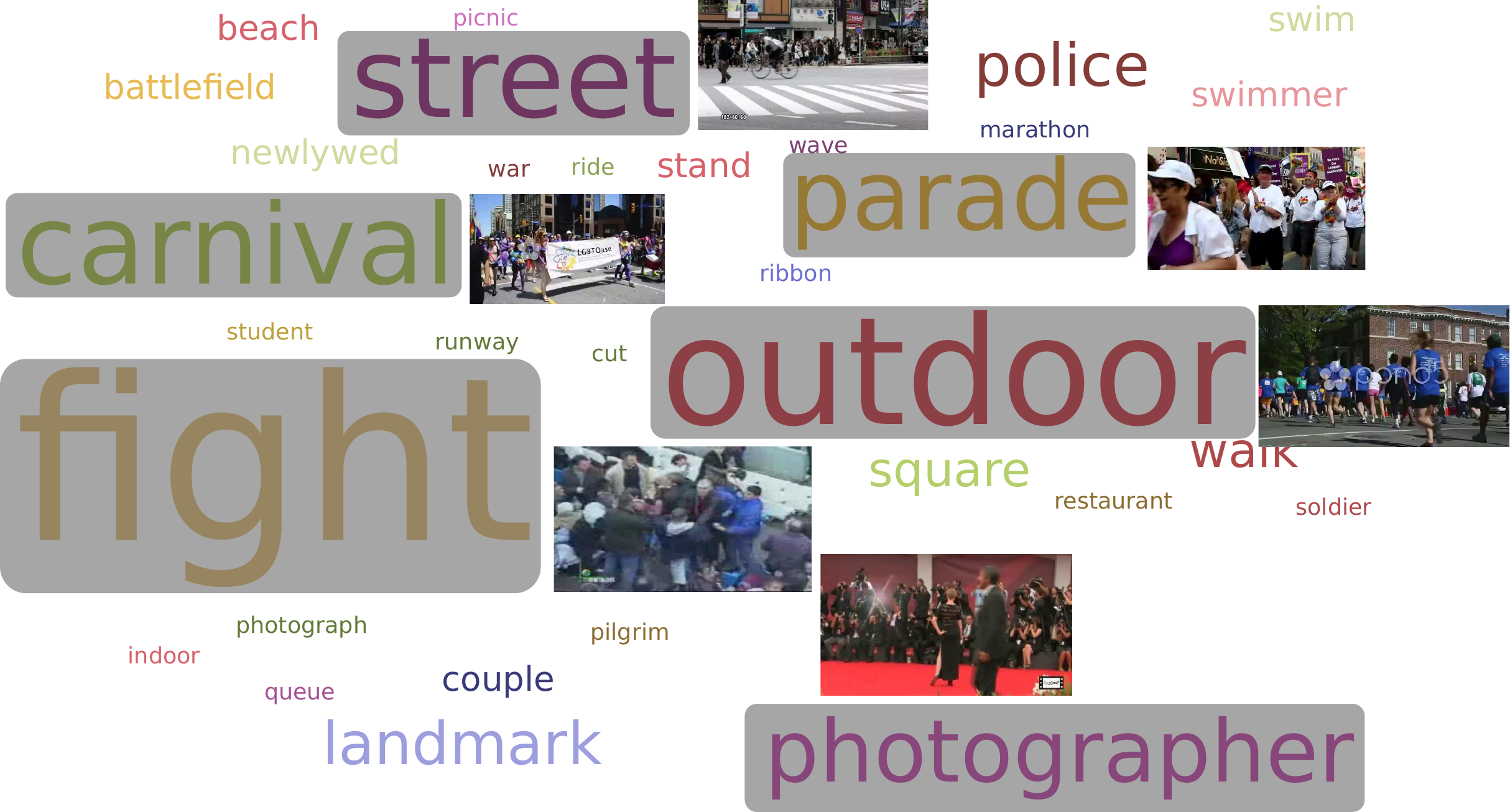}}\\
\subfloat[``marathon'' as novel attribute]{\includegraphics[width=0.75\linewidth]{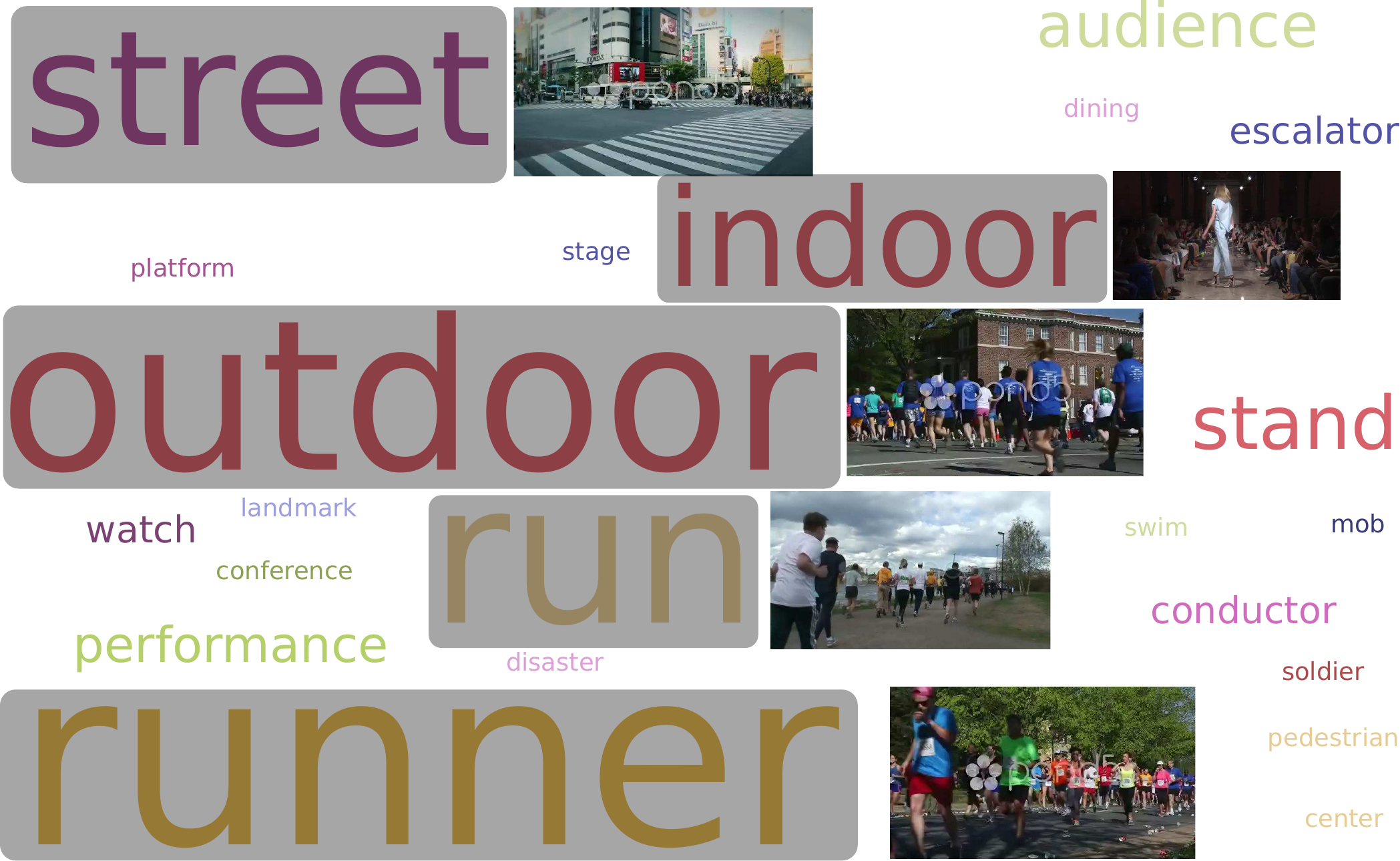}}
\caption{Importantce of known attributes w.r.t. novel event/attributes. The fontsize of each attributes is proportional to the conditional probability e.g. $p(``violence"|y_p)$.}\label{fig:WordCloud}
\end{figure}

Recall that the key step in our method's approach to zero-shot prediction is to estimate the visual co-occurrence (between known attributes and held out zero-shot attributes) based on the textually derived word-vectors of each attributes. To illustrate what is learned, we visualize the predicted importance of 94 attributes from WWW in terms of supporting the detection of the held out attribute  ``\textit{violence}''. The results are presented as a word cloud in Fig.~\ref{fig:WordCloud}, where the size of each word/attribute $p$ is proportional to the conditional probability e.g. $p(``violence"|y_p)$. As we see from Fig~\ref{fig:WordCloud}(a), attribute - ``\textit{fight}'' is the most prominent attribute supporting the detection of ``\textit{violence}''. Besides this, actions like ``\textit{street}'', ``\textit{outdoor}'' and ``\textit{wave}'' all support the existence of ``\textit{violence}'', while `\textit{disaster}' and `\textit{dining}' among others do not. We also illustrate the support of ``mob'' and ``marathon'' in Fig~\ref{fig:WordCloud}(b) and (c) respectively. All these give us very reasonable importance of known attributes in supporting the recognition of novel attributes.

\section{Conclusions}

Crowd behaviour analysis has long been a key topic in computer vision research. Supervised approaches have been proposed recently. But these require  exhaustively obtaining and annotating examples of each semantic attribute, preventing this strategy from scaling up to ever expanding dataset sizes and variety of attributes. 
Therefore it is worthwhile to develop recognizers that require little or no annotated training examples for the attribute/event of interest. We address this by proposing a zero-shot learning strategy in which recognizers for novel attributes are built without corresponding training data. This is achieved by learning the recognizers for known labelled attributes. For testing data, the confidence of belonging to known attributes then supports the recognition of novel ones via attribute relation. We propose to model this relation from the co-occurrence context provided by known attributes and word-vector embeddings of the attribute names from text corpora. 
Experiments on zero-shot multi-label crowd attribute prediction prove the feasibility of zero-shot crowd analysis and demonstrate the effectiveness of learning contextual co-occurrence. A proof of concept case study on transfer zero-shot violence recognition further demonstrates the practical value of our zero-shot learning approach , and its superior efficacy compared to even fully supervised learning approaches.
\bibliographystyle{plain}
\bibliography{./chapter}

\end{document}